\documentclass{article}
\usepackage[T1]{fontenc}
\usepackage{amssymb}
\usepackage{amsmath}
\usepackage{amsthm}
\usepackage[a4paper]{geometry}
\usepackage{multirow}

\usepackage{algorithm}
\usepackage{algorithmic}
\usepackage[round]{natbib}
\usepackage{authblk}       % For author affiliations

\usepackage{graphicx}

\usepackage{tabularx}

\usepackage{booktabs}
\usepackage[table]{xcolor}

\usepackage{caption}
\usepackage{subcaption}
\usepackage{array}
\usepackage{longtable}
\usepackage{enumitem}
\usepackage{booktabs}
\usepackage{adjustbox}

\definecolor{cvprblue}{rgb}{0.21,0.49,0.74}
\usepackage{hyperref}
\hypersetup{colorlinks,allcolors=cvprblue,unicode=true}

\newcommand\blfootnote[1]{%
  \begingroup
  \renewcommand\thefootnote{}\footnote{#1}%
  \addtocounter{footnote}{-1}%
  \endgroup
}

\title{Do Vision Models Develop \\Human-Like Progressive Difficulty Understanding?}

\author{
  Zeyi Huang$^{1\star}$, Utkarsh Ojha$^{1\star}$, Yuyang Ji$^2$, Donghyun Lee$^1$, 
  Yong Jae Lee$^{1}$ \vspace{0.8em}
  \\
  {\hspace{-1.0cm}$^1$University of Wisconsin-Madison \hspace{0.9cm} $^2$UIUC}\\
}

\date{}

\begin{document}
\maketitle

\begin{abstract}
When a human undertakes a test, their responses likely follow a pattern: if they answered an easy question $(2 \times 3)$ incorrectly, they would likely answer a more difficult one $(2 \times 3 \times 4)$ incorrectly; and if they answered a difficult question correctly, they would likely answer the easy one correctly. Anything else hints at memorization. Do current visual recognition models exhibit a similarly structured learning capacity? In this work, we consider the task of image classification and study if those models' responses follow that pattern. Since real images aren't labeled with difficulty, we first create a dataset of 100 categories, 10 attributes, and 3 difficulty levels using recent generative models: for each category (e.g., dog) and attribute (e.g., occlusion), we generate images of increasing difficulty (e.g., a dog without occlusion, a dog only partly visible). We find that most of the models do in fact behave similarly to the aforementioned pattern around 80-90\% of the time. Using this property, we then explore a new way to evaluate those models. Instead of testing the model on every possible test image, we create an adaptive test akin to GRE, in which the model's performance on the current round of images determines the test images in the next round. This allows the model to skip over questions too easy/hard for itself, and helps us get its overall performance in fewer steps.
\end{abstract}

\section{Introduction}
\label{sec:intro}
\blfootnote{$^\star$ equal contribution.}

\begin{figure*}[t]
    \centering
    %\vspace{-0.7cm}
    \includegraphics[width=.9\textwidth]{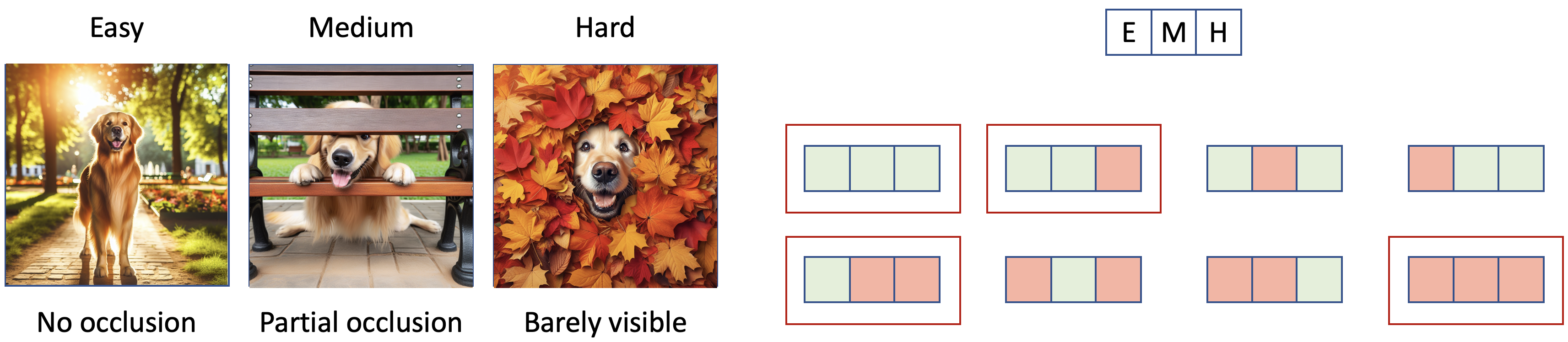}
    \vspace{-5pt}
    \caption{\textbf{Left:} Sample images from our proposed test set generated using GPT-4 + DALL-E 3. For the class of \textit{golden retriever} and attribute \textit{occlusion}, we generate images of varying difficulty. Intuitively, it is easier to classify the leftmost image as \textit{golden retriever} compared to rightmost image. \textbf{Right:} Possible responses (correct/incorrect) of a model on the easy/medium/hard image on the left side. \textit{Cannot solve easy one $\rightarrow$ cannot solve difficult one. Can solve difficult one $\rightarrow$ can solve easy one}: this hypothesis is only satisfied in 4 (in red) out of 8 possibilities.}
    \label{fig:teaser}
    \vspace{-5pt}
\end{figure*}

Imagine a math teacher grading a student's answer sheet, and finds that they got the answer of $(2 \times 3)$ wrong but the answer for $(2 \times 3 \times 4)$ right. The teacher will rightly wonder whether the student properly learnt the concept of multiplication or whether they memorized the answer to the more difficult question. This is because there is a characteristic way in which humans learn any concept: if they cannot answer an easy question, they very likely \emph{cannot} answer a more difficult one. And conversely, if they can answer a difficult question, they most certainly \emph{can} answer an easier one as well. Neural networks are also trained to learn concepts to perform a task. Do they also learn those concepts in a similarly characteristic way?
%can also be thought of as learning concepts

In this work, we study this question for the task of image classification. Our goal is to see if modern visual recognition systems (e.g., ConvNext~\citep{liu2022convnet}, ViT~\cite{dosovitskiy2020image}) have that human-like behavior to easy/hard-to-classify images. Since there does not exist appropriate real datasets labeled with ground-truth difficulty, we \emph{generate} one instead. Recent image generative models have become capable of generating very high quality images \citep{rombach2022highresolutionimagesynthesislatent}, good enough to be used in training recognition models \citep{yu2023diversify, azizi2023}, and we believe that they are good enough to be used for our evaluation. With the aid of recent large language and generative models (GPT-4~\cite{achiam2023gpt} + DALL-E 3~\cite{ramesh2021zero}), we design a prompting system to generate descriptions of images of three levels of difficulty. For example, \textit{an image of a fully visible dog} and \textit{an image of dog only partly visible} can be considered to be image descriptions of an easy- and hard-to-classify images respectively. We use DALL-E 3 to take in these different difficulty level prompts and generate images while faithfully preserving the desired attributes. Fig.~\ref{fig:teaser} (left) gives an example. 

Once we have the easy, medium and hard-to-classify test images, we record if the model predicts the class correctly or incorrectly. Fig.~\ref{fig:teaser} (right) depicts the 8 possibilities of model's behavior (green/red represent correct/incorrect response). If the model truly learns to classify images by developing the aforementioned notion of easy/hard concepts, then its responses should fall under 4 out of the 8 possibilities highlighted in a red box. Our first key finding is that, for most of the current visual recognition models, their responses \emph{do indeed} fall under the 4 highlighted categories around 80-90\% of the time. This result hints that even without an explicit supervision, visual recognition models learn to learn things in a structured way.

While an intriguing result in its own, we believe that this can have applications, especially in the way we evaluate models. We take inspiration from how students are often tested using standardized tests, like the Graduate Record Examination (GRE), for admissions into U.S.~universities. These tests are adaptive in nature, where questions in the next round depend on how well the student does in the current one. So, for example, if the student cannot solve easy-medium questions, there is not much point giving them difficult questions in the next round; i.e., one can reliably \emph{predict} that they will get zero points for those hard questions. We develop a similar GRE-type test to evaluate visual recognition models on the generated dataset proposed above. The test is broken up into multiple rounds. In the first round, the model is shown images of medium difficulty on average. Its score in this round determines the distribution of easy/medium/hard questions in the next round. That is, similar to GRE, we can skip over images that are too easy/hard for the model to classify. Thus, instead of evaluating the model on every possible image in the test set, this way of dynamically selecting the images helps approximate that total score of the model on the whole set using only 25\% of the test images. 

Additionally, the newly proposed dataset can have usefulness in and of itself. We generate images from 100 categories taken from ImageNet~\citep{deng2009imagenet}. For each category, we consider 10 attributes. Within each attribute, we generate 12 images for 3 levels of difficulty, bringing the total number of images to 36,000. However, different from standard benchmarks like the ImageNet validation set, these 36k images are labeled with attribute value, difficulty, in addition to the ground-truth class. This can enable analysing models on a much finer level (e.g., ResNet-50 struggles to detect dogs from a side view). 

In summary, our work has the following contributions. We present a new method to study the learning dynamics of modern visual recognition systems using the concept of example difficulty. To do this, we create a new test set of synthetic images labeled with class, attribute, and difficulty level. Our results indicate that most of the models do in fact develop a semantically meaningful notion of example difficulty while learning visual concepts, without having access to any external supervision. Using this newly found property, we develop a multi-round adaptive test, inspired by GRE, which steers the future test images according to a model's ongoing performance. This facilitates skipping over too easy/hard questions, and helps assess a model's performance using a fraction of test images.

\section{Related Work}
\label{sec:formatting}

\paragraph{Neural network learning mechanisms.}
Understanding how neural networks learn concepts has been a key focus in deep learning research. Some have studied neural networks' intriguing tendency to generalize vs memorize (e.g., random labels) on the training data \citep{zhang2021understanding, arpit2017closer}. Along this line, some studies suggest that the generalization ability depends more on training data than model capacity \citep{dinh2017sharp, krueger2017deep}. A different angle of understanding how neural networks function is via important feature visualization e.g., visualizing partial derivatives \citep{simonyan2013deep} and Class Activation Mapping (CAM) \citep{zhou2016learning} highlight important image regions used by CNNs for decisions. Others study how the training process incentivizes neural networks to focus more on learning certain concepts, and implicitly ignoring others. Neural networks often prioritize easy-to-learn features like texture \citep{geirhos2018imagenet}, while harder-to-learn samples, such as shapes, may be neglected \citep{geirhos2020shortcut}. Minority samples in datasets can also be overlooked \citep{mehrabi2021survey}. The idea that there might exist some notion of easy and difficult-to-learn concepts has been used to define a curriculum for training neural networks, where models are trained on tasks of increasing complexity with explicit or implicit guidance~\citep{bengio2009curriculum, saxena_curriculum}. However, merely defining such a training curriculum will not guarantee that we get a model having the characteristic described in Fig.~\ref{fig:teaser}. This is because (i) curriculum learning doesn't explicitly enforce the principle that a model can learn harder concepts \emph{only} if they correctly learn the easier one first; and (ii) there is a known problem of forgetting in neural networks, where they might even forget easier concepts learned earlier but still correctly learn more difficult ones later. To the best of our knowledge, we are the first to investigate whether such property \emph{emerges} in neural networks automatically \emph{after} their complete training.

\vspace{-10pt}
\paragraph{Datasets for studying models' properties.} The standard way to evaluate image classifiers is to obtain its accuracy on a human-collected test set. However, there have been concerns that some test sets, e.g., ImageNet~\citep{deng2009imagenet}, have become saturated~\citep{mayilvahanan2023does}. To address this, several works have proposed new datasets to test the robustness of models to distribution shifts~\citep{recht2019imagenet,wang2019learning, barbu2019objectnet, hendrycks2021many, taesiri2024imagenet, hendrycks2019benchmarking, geirhos2018imagenet}. Others use synthetic data to test visual recognition models. Photorealistic Unreal Graphics (PUG) dataset~\citep{bordes2024pug}, generated using Unreal Engine, provides photorealistic, controllable synthetic data and can test a model's robustness to factors like pose, background, size, texture, and lighting. ImageNet-D~\citep{zhang2024imagenet} uses Stable Diffusion to create challenging images, while Spawrious~\citep{lynch2023spawrious} leverages it to generate datasets with spurious correlations. There are also datasets which measure some notion of difficulty associated with an image, label pair. Some have defined difficulty on a object class level~\citep{barbu2019objectnet}. Others define difficulty on an individual image level; e.g., using the time humans need to correctly classify images~\citep{mayo_difficulty}, whether the image gets classified correctly/incorrectly by most models~\citep{meding_difficulty}, or based on on its expected accuracy given training sets of varying size~\citep{jiang_difficulty}. However, all these notions of difficulty scores are isolated, and not anchored to any understandable attribute. We believe that an intuitive way to understand what makes an image difficult is to imagine what would have made it easy to classify. The third image in Fig.~\ref{fig:teaser} is difficult to classify as dog because it is occluded. Here, occlusion becomes a sample attribute along which we can make an image easy or difficult. We present a method to create a dataset using modern generative models which has these difficulty labels.

\vspace{-10pt}
\paragraph{Adaptive model evaluation.}
Inspired by computerized adaptive testing (CAT)~\citep{van2000computerized}, used in exams like the GRE, we develop algorithms for adaptive testing in image classification benchmarks. Similar to CAT, which does not require all questions to be answered and instead efficiently assesses examinees with fewer questions based on examinee responses, our framework evaluates classification models by selecting a subset of samples, reducing computational demands while maintaining assessment accuracy. \cite{prabhu2024lifelong} presents an efficient evaluation framework to maintain a lifelong benchmark by leveraging dynamic programming to rank and sub-select samples, whereas our approach generates unseen images using DALL-E, ensuring models are tested on genuinely new data, mitigating the risk of data leakage, and enhancing the reliability of generalization evaluation.information within the scene.

\section{Models' Behavior on Easy \texorpdfstring{$\rightarrow$} Hard Images}

Humans learn concepts in a progressive way - it is only possible for them to solve hard questions if they have the ability to solve easier ones first \citep{Zacks2001EventSI, Newtson1973AttributionAT}. We want to study if image classification models also learn concepts in a similar way. To do this, we first explain in Sec.~\ref{sec:gen_data} our process of creating a test data having images of various difficulty (akin to questions for humans). Then, in Sec.~\ref{sec:progression_score}, we present a method to analyze a model's response to such images to see if it mimics a human-like way of learning concepts. 

\begin{figure}[t]
    \centering
    %\vspace{-0.7cm}
    % \includegraphics[width=1\textwidth,page=2]{figures/tesear.pdf}
    \includegraphics[width=0.9\textwidth]{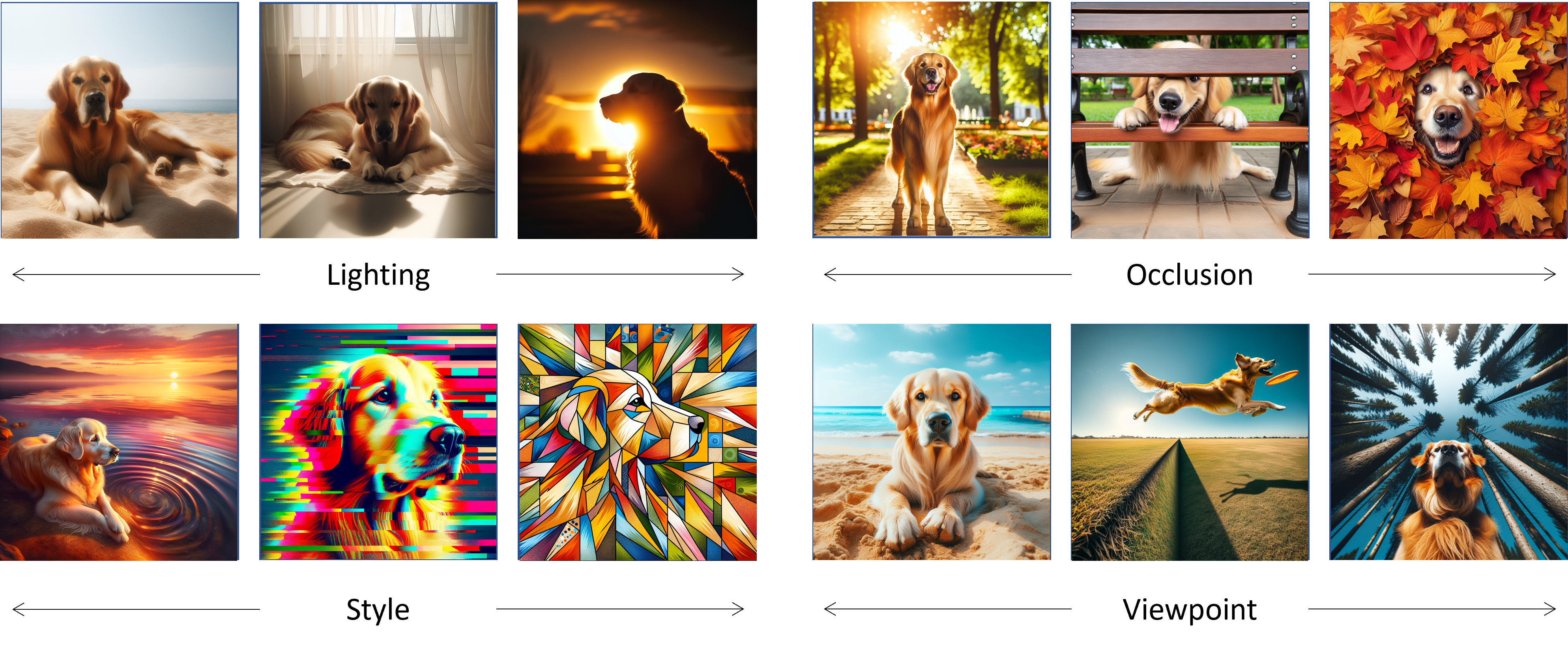}
    \vspace{-0.5cm}
    \caption{\textbf{Visualizing the difficulty of test samples.} All of the images are generated using our proposed pipeline. In each quadrant, we focus on one attribute (e.g., lighting, in the top left), and from left to right we show the images becoming progressively more difficult to be classified correctly.}
    \label{fig:difficulty_examples}
    \vspace{-0.1in}
\end{figure}

\begin{figure*}[t]
    \centering
    \includegraphics[width=.9\textwidth]{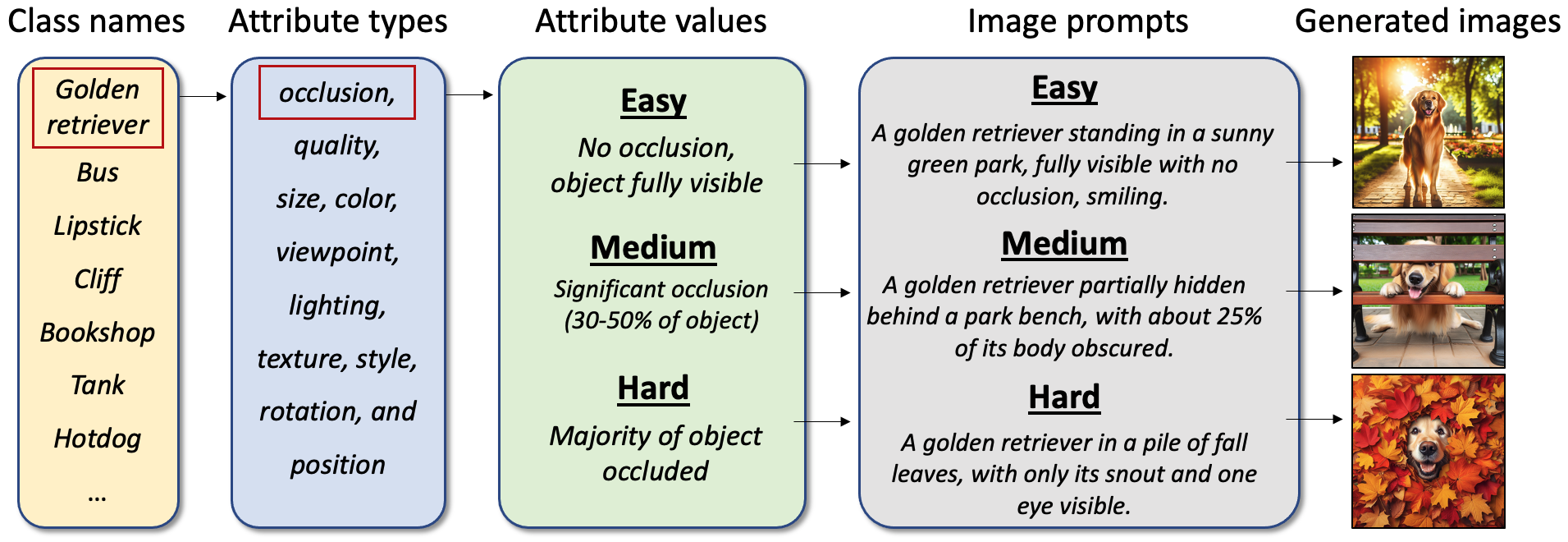}
    \vspace{-0.25cm}
    \caption{\textbf{Overview of the test set generation process.} The first step is to collect the names of the image categories that we wish to test the models on. We then prompt GPT-4 to generate the appropriate attribute values for those categories with various levels of difficulty. Using those, we again prompt GPT-4 to generate text prompts for a category (golden retriever), attribute (heavy occlusion) combination. Finally, we use DALL-E 3 to generate the corresponding images.}
    \label{fig:gen_process}
    \vspace{-0.1in}
\end{figure*}

\subsection{Dataset Creation}\label{sec:gen_data}
For the task of image classification, the standard data format is images paired with their ground truth labels $\mathcal{D} = \{(x_1, y_1), (x_2, y_2), ...\}$. However, for our purpose, not only do we need the information that $y_i$ is the ground-truth label of $x_i$, but we also need to know how \emph{difficult} it is for $x_i$ to be classified as $y_i$. Hence, our first task is to formalize this notion of difficulty for our problem.

\subsubsection{Understanding sample difficulty} 
Consider Fig.~\ref{fig:difficulty_examples} and observe the group of three images in the top right corner. All of the images have `golden retriever' as the main subject. Yet, we, \emph{as humans}, can intuit that it is easiest to correctly classify the content of the left most image and most difficult to do it for the right most one. This is because from left to right, the dog is getting more occluded. Other triplets show similar easy $\rightarrow$ hard progression for other attributes. The important point from these figures is that the difficulty of a sample is best understood in relation to a particular attribute type (e.g., occlusion). However, to our knowledge, no dataset of real images contains human annotations indicating difficulty about a sample in this way.

Hence, given the recent advances in language and image generative models, we propose to \emph{synthesize} images with the desired attributes. Text-to-image generative models can now generate very high quality images \citep{ rombach2022highresolutionimagesynthesislatent, openai2024dalle3}, so good that they have even been used to train image classification models and shown promise \citep{yu2023diversify, azizi2023}. Furthermore, using the text medium, we can use very precise language to describe more easy/hard-to-classify images. Hence, we propose to use these generative models to design an evaluation setup.

\subsubsection{Overall image generation pipeline}
Since we want to generate images of varying difficulty using text, we need the following information for a particular prompt: (i) class name (e.g., golden retriever), (ii) attribute type (e.g., occlusion), and (iii) difficulty type (e.g., hard) for that attribute. Using (ii) and (iii), we can specify a particular attribute value. Here is an example: \textit{An image of a golden retriever heavily occluded by a door}. In this case, \textit{golden retriever} is the image class, \textit{heavily occluded} is the \textit{hard} difficulty for the occlusion attribute. The first step is to collect the names of the classes we wish to evaluate on. We use 100 object categories out of 1000 classes in ImageNet (see appendix \ref{appendix:categories} for the complete list). The next step is to generate the attribute values, similar to \textit{heavily occluded} in the above example. 

\textbf{Generating the attributes.} We first prompt GPT-4~\citep{achiam2023gpt} to list 10 common attribute types that can help describe image content. The second column of Fig.~\ref{fig:gen_process} shows the list. After this, we again prompt GPT-4 with the following for each of those attributes:

\textsf{``To generate text prompts for DALL-E that will generate images of varying difficulty levels for vision models to classify, please create nine levels of difficulty based on <attribut name> attributes and group the nine levels of difficulty into categories of easy, medium, and hard.'' }

To this, GPT-4 responds with a list of difficulty varying attribute values (descriptions) for each attribute type. As an example, here is how the difficulty of an image can vary along the attribute of \textit{occlusion} - (i) Easy: ``No occlusion, object fully visible''; (ii) Medium: ``Significant occlusion (30-50\% of object)''; (iii) Hard: ``Majority of object occluded (70-90\%)''.  

\textbf{Generating text prompts.} Given these basic units, we prompt GPT-4 one last time to generate the text descriptions for DALL-E 3 to generate the images. This description is produced by combining the information about a certain class with a certain attribute value; e.g., \textit{golden retriever} is combined with \textit{heavy occlusion} to produce the prompt - \textit{``A golden retriever in a pile of fall leaves, with only its snout and one eye visible''}. Fig.~\ref{fig:gen_process} details the whole process. For detailed prompts used throughout, refer to Appendix \ref{sec:prompt_design}.

\textbf{Dataset size:} There are a total of 100 classes, each having the same 10 attributes. For each attribute, there are a total of 3 difficulty levels. And for each difficulty level, there are 12 images. Hence, the total size of our dataset is $10 \times 100 \times 3 \times 12 = 36000$ images. This dataset is balanced across types of class, attribute, and difficulty level.

\begin{figure}[t]
    \centering
    %\vspace{-0.7cm}    
    \includegraphics[width=.9\textwidth]{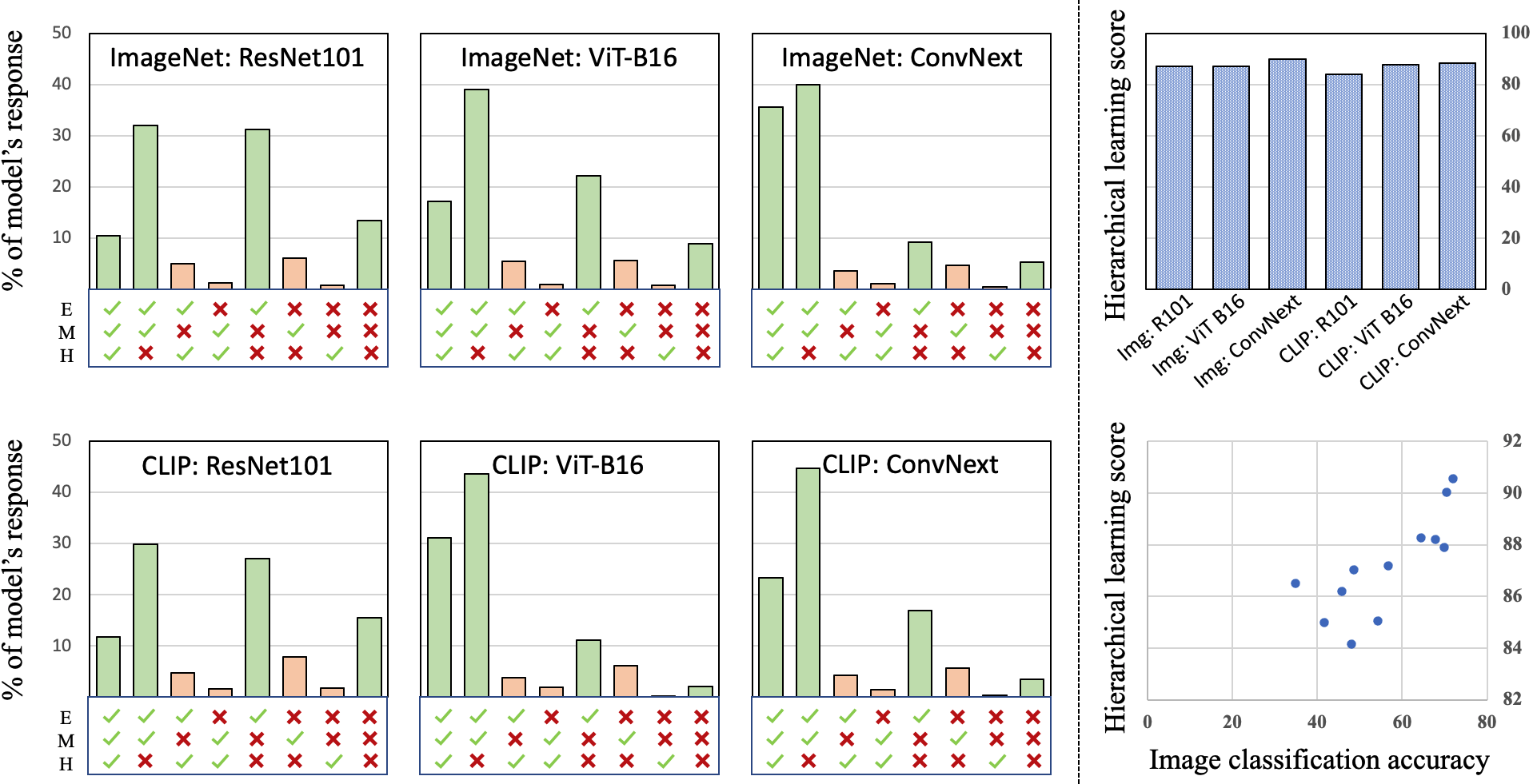}
    \caption{\textbf{Left:} Plots depicting \% of model's behavior on 12k triplets over the 8 possible patterns of \textbf{E}asy, \textbf{M}edium, \textbf{H}ard. The bars corresponding to principle-following pattern are colored green; others, red. All models behave according to the hierarchical learning principle. \textbf{Top right:} Hierarchical learning score of 6 vision models. Most achieve a score higher than $85\%$. \textbf{Bottom right:} Scatter plot of top-1 accuracy on our test set vs hierarchical learning score of 12 models. PCC value is 0.77.}
    \label{fig:hscore_results}
    \vspace{-0.1in}
\end{figure}

\subsection{Hierarchical learning score}\label{sec:progression_score}

We now describe how to use this dataset of easy/medium/hard-to-classify images to study whether models learn concepts hierarchically. If we consider the combination of each class with its attribute type, there will be a total of $100 \times 10 = 1000$ combinations (\{\textit{Golden retriever, occlusion}\} is one such example). For each such class-attribute combination, we create 12 triplets, each having one easy, one medium, and one hard image. Hence, there are a total of 12,000 triplets of test images. 

% that is of interest to us
For each visual recognition model, we test it on these images and record whether their prediction matches the ground-truth object category. The model's response on each triplet will follow one of 8 patterns depending on whether it gets the easy/medium/hard image correctly or incorrectly classified; these are shown in Fig.~\ref{fig:teaser} right (red/green colors mean correct/incorrect prediction respectively). We compute the percentage of each of these 8 patterns over all 12,000 triplets. Now, if we consider the following human-like hierarchical learning principle - \textit{A model should correctly answer a harder question only if it can answer all easier questions}, we see that only 4 out of those 8 patterns satisfy the requirement (highlighted with red box). We call them `principle-following patterns', and define the \textbf{hierarchical-learning score} to be the \emph{percentage of model's responses on 12,000 triplets that fall under principle-following patterns}. The more a model follows this principle, the higher its hierarchical-learning score.% should be

\textbf{Experiments and Results:} We choose three popular architectures - (i) ViT-B16~\citep{dosovitskiy2020image}, (ii) ConvNext~\citep{liu2022convnet} and (iii) ResNet-101~\citep{he2016deep} - each trained on ImageNet1k~\citep{deng2009imagenet} and LAION~\citep{schuhmann2022laion} using cross-entropy and CLIP objective~\citep{radford2021learning}, respectively. This gives us a total of 6 trained models which we test using the setting described above. The results are shown in Fig.~\ref{fig:hscore_results} (top), where we plot the percentage of a model's behavior on 12,000 triplets over all the patterns. We color the bars corresponding to principle-following and not following patterns as green and red respectively. Deriving from these plots, we also report the hierarchical-learning score of these 6 models in Fig.~\ref{fig:hscore_results} (bottom left). First, for all models, the majority behavior on triplets falls under the principle-following pattern, and almost all get a high hierarchical-learning score of $>$85\%. In fact, in all cases, the 4 most frequent behaviors correspond to those 4 principle-following patterns. The most frequent behavior across all models is \{Easy: \checkmark, Medium: \checkmark, Hard: $\times$ \}, while the least frequent one is \{Easy: $\times$, Medium: $\times$, Hard: \checkmark \}. Also, whether a model was trained on ImageNet using cross entropy loss or on LAION to align images and text, e.g., ResNet101 vs CLIP:ResNet101, the behavior remains roughly the same. These results indicate that common visual recognition models do in fact follow the human-like hierarchical learning principle, even when there is no explicit supervision to do so!  

\textbf{Model's hierarchical learning score vs accuracy:} Consider two extreme scenarios. Models A and B are so good and bad at classifying images that most of their behavior is of the form \{\checkmark, \checkmark, \checkmark \} and \{$\times, \times, \times$\} respectively. Even though their top-1 classification accuracies could be vastly different, they would both still have a very high hierarchical-learning score. So, it is unclear what, if anything, the hierarchical-learning score has to do with classification accuracy. We empirically study this, by collecting a total of 12 models (6 additional ones compared to previous study; see appendix) and for each, compute its hierarchical-learning score and top-1 accuracy on our test data. The scatter plot in Fig.~\ref{fig:hscore_results} (bottom right) shows these data points. There is a correlation between the top-1 accuracy and the hierarchical learning score; the Pearson correlation coefficient is 0.77. Since this is merely a correlation, it is difficult to say whether more accurate models become more accurate \emph{because} they learn to learn concepts in a human-like style. Nevertheless, a positive correlation is a sign that, contrary to the extreme example we discussed above, the learning dynamics of very incapable and capable models is not symmetric (i.e., when a model is good/bad accuracy-wise, it is also good/bad in the hierarchical-learning sense). However, we point out that even the least hierarchical-learning score for a model is 84.2, which, for the purposes of this work, we consider high enough to say that the corresponding model still follows the hierarchical-learning principle.

\begin{figure}[t]
\centering
\includegraphics[width=0.8\textwidth]{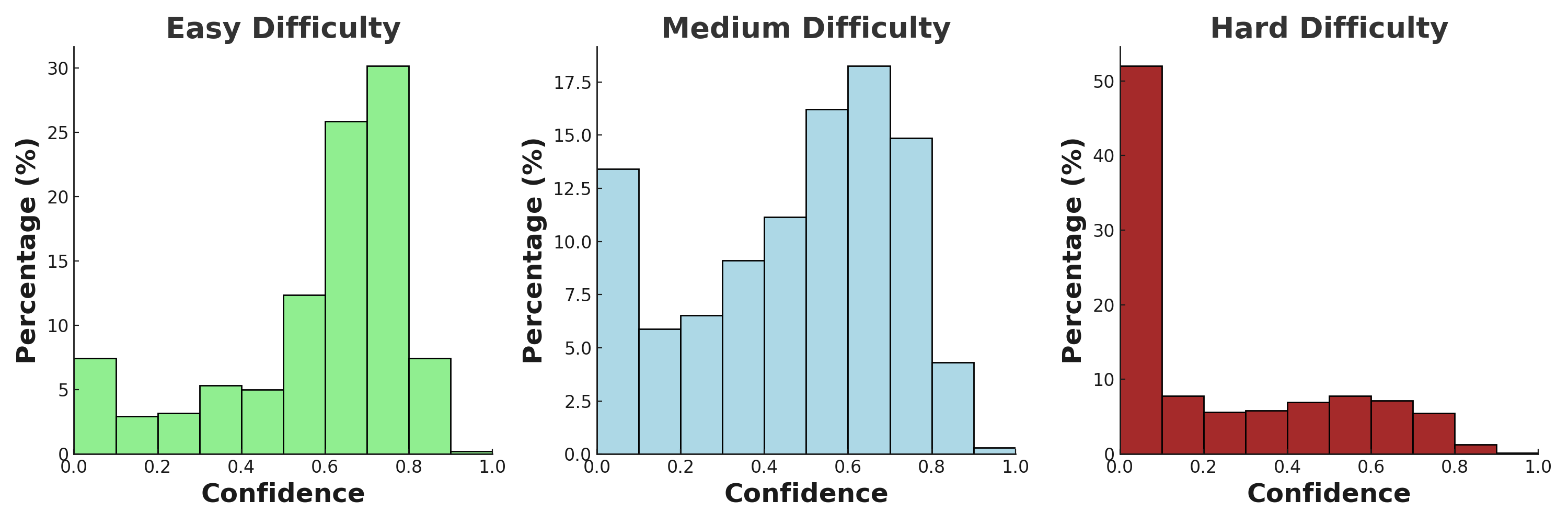}
\caption{\% of our dataset grouped according to classification confidence for the Easy, Medium, and Hard difficulty levels. We average the sample numbers across six selected classifiers (ViT-B16~\citep{dosovitskiy2020image}, ConvNext~\citep{liu2022convnet}, ResNet-101~\citep{he2016deep}, trained on ImageNet1k~\citep{deng2009imagenet} and LAION~\citep{schuhmann2022laion}). See Appendix for more confidence visualization of different classifiers.}
\label{fig:synthetic}
\vspace{-0.2in}
\end{figure}

\subsubsection{Evaluating correctness of difficulty levels}% in generated images

A key ingredient to test our hypothesis, i.e., whether vision models follow the hierarchical learning principle, has been the usage of generated images. The images shown in Fig.~\ref{fig:difficulty_examples} do look good to our eyes and seem to make sense (some are easier to classify, others more difficult). However, to make sure that our dataset as a whole is meaningful and does not contain garbage images, we verify whether those same images of various difficulty can be appropriately categorized into their respective difficulty levels. To accomplish this, we use six classifiers to assess the prediction confidence (probability after softmax) for each sample at each difficulty level. For the easy difficulty level, we expect a high proportion of samples to exhibit high prediction confidence. Conversely, for the hard level, we anticipate a significant number of samples to display low prediction confidence. For the medium level, we expect the prediction confidence of many samples to fall between the easy and hard difficulty levels. By analyzing the distribution of prediction confidence across the difficulty levels for our entire dataset of 36,000 images in Figure~\ref{fig:synthetic}, we validate the efficacy of our image generation process and ensure that the generated images accurately represent the intended difficulty levels.

In addition, we conducted a user study to evaluate the alignment between difficulty levels of generated images (Easy, Medium, Hard) and human perception using pairwise comparisons. We sampled 900 images from 10 classes, 10 attributes per class, and 3 difficulty levels per attribute. 10 participants evaluated 2 classes each, completing 540 comparisons per participant. Participants were shown two images at a time and asked to select the more difficult one. Each image appeared in multiple comparisons to ensure robustness. Human responses were used to fit a Bradley-Terry Model, generating a continuous difficulty score for each image. The computed correlation values are $r = 0.871$ (Pearson), $\rho = 0.883$ (Spearman), and $\tau = 0.749$ (Kendall’s Tau), indicating strong alignment between difficulty labels of generated images and human perception. Please see Appendix Sec~\ref{sec:user_stduy} for more details.

\section{Adaptive Testing of Image Classifiers}\label{sec:adapt_test}

The fact that humans learn concepts in a hierarchical way has had applications in the way their learning gets tested. Many standardized tests like the GRE are adaptive in nature. Based on how well the student is doing at any point, future questions are adapted to match the student's ability. For example, if a student is struggling to answer questions like $(2 \times 3)$ and $(4 \times 5)$, there is not much point in testing them on $(2^2 + 3^2) \times (4 + 5)$. We can reasonably predict that they will get zero points for the latter question, and hence can avoid testing them on all possible (easy/medium/hard) questions, while still being able to accurately assess a student's capability. Similarly, our setting also has a test set of 36k images uniformly spread over 100 object categories, 10 attribute types, and 3 levels of difficulty. Given that visual recognition models also seem to learn in a hierarchical way, we investigate a similar GRE-type test for them, using which we can get a good estimate of model's overall performance without needing to test it on every single easy/medium/hard image. 

\begin{figure}[t]
    \centering
    %\vspace{-0.7cm}
    % \includegraphics[width=1\textwidth,page=2]{figures/tesear.pdf}
    \includegraphics[width=.9\textwidth]{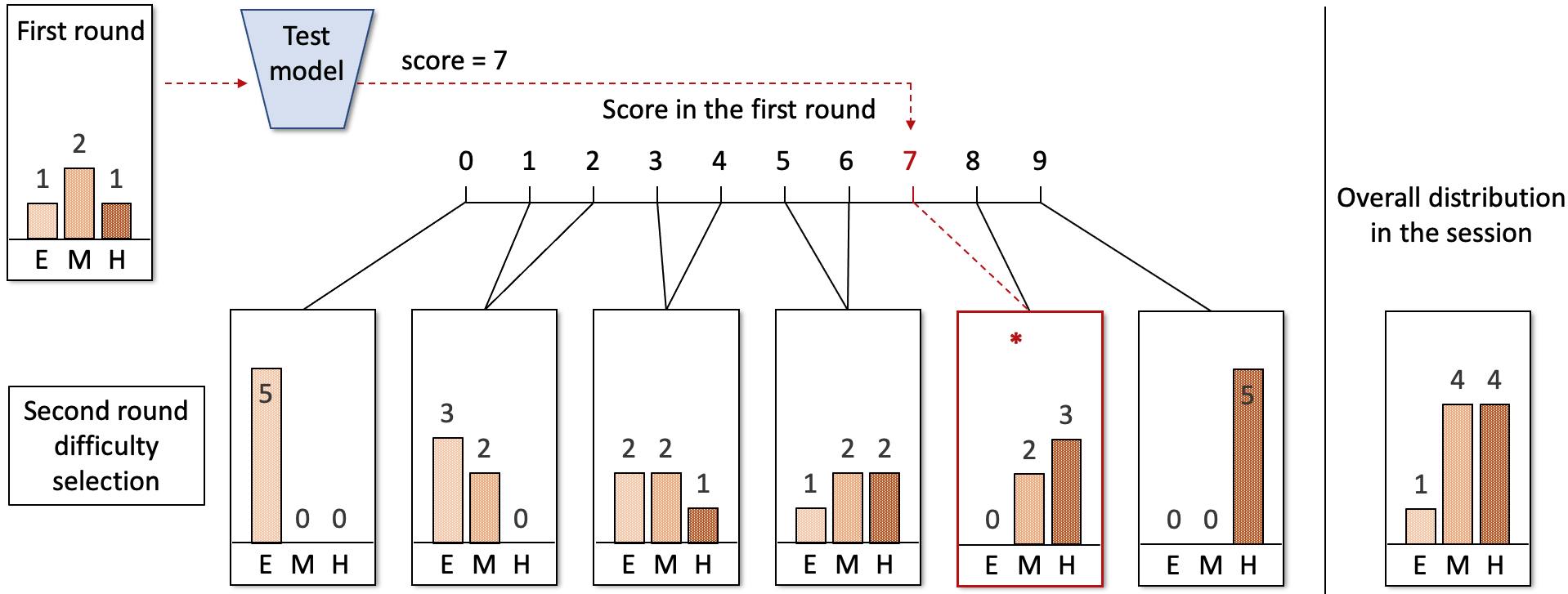}
    \caption{\textbf{Adaptive testing of a classifier.} The test involves two rounds. Similar to GRE, the first round is of, on average, medium level difficulty, consisting of 4 test images (1 easy, 2 medium, 1 hard). The model gets a score (max = 9, min = 0) based on which the distribution of images for the next round is chosen. We show an example of a model getting a score of 7 in round 1, because of which in next round there are 0 easy, 2 medium, and 3 hard images. \textbf{Right:} Throughout the whole session, the model gets tested on a total of 9 images; in this case, 1 easy, 4 medium and 4 hard images.}
    \label{fig:adaptive_test}
    \vspace{-0.1in}
\end{figure}

To test a model, we start by iterating over all the classes and attribute types. In each such class + attribute type combination (there are 1000 of these), there are 3 difficulty levels, each with 12 images, i.e., $3 \times 12 = 36$ test images. The goal is to predict how a model performs on these 36 images without needing to evaluate it on all 36 images. Hence, we conduct a test borrowing principles from GRE. Instead of the model being tested on all images in the same round at once, we consider two rounds of test images. In the first round, we randomly select 1 easy, 2 medium, and 1 hard image. The average difficulty is medium. The model is tested on these 4 images, and we assign it a score based on how well it did. Similar to GRE, one gets more points for solving harder images. We assign the model 1, 2, and 4 points for correctly classifying an easy, medium, and hard image, respectively; and 0 points for misclassification. In the next round, the model is tested on 5 new images. The distribution of these new images depends on the model's total score in the first round (min/max score is 0/9 respectively). The mapping from total score to second round images' distribution is shown in Fig.~\ref{fig:adaptive_test}. Similar to the first round, the model gets assigned points for correctly classifying images. All throughout the process, the model is tested only on $x$ easy, $y$ medium, and $z$ hard test images, where $x + y + z = 9$. So, after the end of the second round, we can collect the following information about the model: (i) its total score, and (ii) classification accuracies on easy/medium/hard images. While our goal is not to suggest that this is the precise way that all models should be tested, we have some practical reasons for choosing this configuration. Please see the appendix Sec~\ref{sec:hyper} for a discussion.

Once the model has gone through all the class/attribute combinations, and we have collected these information from each session (first + second round), there are multiple ways to combine that information to analyze the model. We can get an attribute level score by averaging each session's score across all the 100 classes. Similarly, we can get attribute level accuracy, either overall across easy/medium/hard questions or per difficulty, also averaged across the classes. This is helpful if we want to study which attributes a model struggles at, discussed more in Sec.~\ref{sec:exp}. Obviously, one can get a global score/accuracy, averaged across attributes dimension, for the model. The key part in all of these analyses is that they can be done using only a fraction of the 36k images. 

The principle of this adaptive test is applicable to all forms of AI models, even to the now popular breeds which are actually taking and excelling at GRE-type exams—large language models which produce a long chain of thought during testing. For such models, like OpenAI o1~\citep{openai_o1}, it might become more cost-effective to evaluate them using a similar adaptive testing. We leave experiments with LLM to future work, and focus on adaptively testing vision models as a proof of concept. 

In the next sections, we discuss how accurate our predictions (score/accuracy) can be when compared against those same values computed over the whole set.

\subsection{Is the test different for different models?}
\vspace{-1pt}

The purpose of the adaptive test is to test more capable models on harder samples, and not-so-capable models on easier ones. Given that in each session (first + second rounds) we test any model on a total of 9 questions, we should hence expect some diversity in what those 9 questions look, on average, for different models. In Table~\ref{table:emh_number}, we plot the average distribution of each/medium/hard test images that 7 visual recognition models face in a session. We see that most models get tested the most on medium questions. However, there are significant differences among certain models. For example, ResNet18, which is the weakest model in the list (ImageNet validation top-1 accuracy = 69.76\%) gets tested much more on easier images (3.51) than hard ones (1.55). In contrast, a stronger \& capable model like ConvNext (ImageNet validation top-1 accuracy = 84.06\%) gets tested much more on harder (3.98) than easier images (1.55). So, these results tells us how different models, owing to their differing capabilities, create their own unique trajectories of test questions that they get evaluated on.  

\begin{table}[t]

\caption{Average number of questions tested per difficulty level for adaptive testing. Models with better performance tend to receive a higher proportion of medium and hard questions, and vice versa.}
\vspace{-3pt}
\label{table:emh_number}
\footnotesize
\centering
\begin{tabular}{l c c c }
\toprule
Difficulty & Easy & Medium & Hard  \\
\midrule
 ResNet18 &3.51 & 3.83 &1.55 \\
 ResNet101& 2.56&3.96&2.48 \\
 ViT-B16& 2.0 &3.89 & 3.07 \\
 ConvNext-B& 1.55& 3.46 & 3.98  \\
 CLIP-RN101& 2.60 & 3.89 & 2.51 \\
 CLIP-ViT-B16& 1.47 & 3.61 & 3.93\\
 CLIP-ConvNext-B& 1.62& 3.79& 3.58\\
 % \midrule
 % Easy extreme case& 6 & 2 & 1\\
 % Hard extreme case& 1 & 2 & 6\\
% \midrule
%  &  &  &  &  &  &   \\
\bottomrule
\end{tabular}
\vspace{-4pt}
\end{table}

\subsection{How closely does Adaptive Evaluation follow Full Evaluation?}
\vspace{-1pt}

We next evaluate our GRE-style adaptive evaluation comparing to what the performance would be with the standard way of evaluating a model on the entire dataset across the ten attributes and three difficulty levels.

To measure classification performance of a model, in addition to standard classification accuracy, we compute a GRE-style score to reward getting more difficult examples correct: $\text{Score} = (\text{correct}_{\text{easy}} \times 1) + (\text{correct}_{\text{medium}} \times 2) + (\text{correct}_{\text{hard}} \times 4)$,
i.e.,  1, 2, and 4 points for correctly classifying an easy, medium, and hard image; and 0 points for misclassification. Note, one advantage of using a GRE-style score is its ability to differentiate between models with similar accuracy. For example, when two models, such as ConvNext-B and CLIP-ViT-B16, achieve close accuracy scores (70.2 and 69.8, respectively) in Table~\ref{table: compare}, the GRE-style score (59.2 vs. 57.6) can highlight the differences by showing which model is more successful in handling more challenging questions.

Our full generated test set has a total of 36,000 images—comprising 12 images for each combination of 100 classes, 10 attributes, and 3 difficulty levels. We refer to the evaluation results from this complete set as `Static 12', which serves as our ground truth (full evaluation baseline).
To validate our adaptive testing method, we first establish a baseline called `Static 3', which is created by randomly selecting 3 images out of the 12 for each difficulty level. This produces a subset containing 100 (classes) × 10 (attributes) × 3 (difficulty levels) × 3 (images) = 9,000 images.
Our adaptive testing then selects 9 samples across the three difficulty levels for each attribute and class, allowing the number of selected images for each difficulty level to vary. This also results in a subset of 100 (classes) × 10 (attributes) × 9 (images) = 9,000 images, matching the size of the Static 3 subset, but with a different difficulty distribution than that of Static 3. We then evaluate the classifier on these subsets, aiming to achieve results that correlate with the full test set (`Static 12') while using fewer images and potentially maintaining smaller errors compared to the Static 3 strategy.

We compare the performance of Static 3 and our adaptive testing with the full evaluation (Static 12) in Table~\ref{table: compare}. This evaluation is repeated three times, with the average error reported, and the standard deviation provided in appendix~\ref{sec:std}. The results indicate that our adaptive testing provides accurate performance detail estimates using fewer test images than the full set and achieves smaller score and accuracy errors compared to the Static 3 strategy. This confirms that based on the hierarchical learning pattern of common image classification models, it is possible to perform adaptive testing to expedite the evaluation process.

% \begin{table}[t]
% %\vspace{-1em}
% \caption{We evaluate each classifier's score and accuracy using Static 3 and our adaptative method, comparing their 10-dimensional vectors---each element representing the score or accuracy of a specific attribute---against Static 12 using the Mean Squared Error. The evaluation is repeated three times, with average error reported, and Std in the supp. Our method provides accurate performance estimates with fewer test images and smaller errors than Static 3, optimizing the evaluation process. (Static $\#$) represents the number of test images used in each level of difficulty for each attribute of a given class.}
% \vspace{-0.8em}
% \label{table: compare}

% \tabcolsep=0.1cm
% \resizebox{\columnwidth}{!}{
% \centering
% \begin{tabular}{l l| c c c  c c c }
% \toprule
% &Classifier &RN101 &ViT-B16& ConvN-B &C-RN101 &C-ViT-B16& C-ConvN-B\\
% \midrule
% Score&Static 12      & 35.3 & 43.8 & 59.2 & 36.4 & 57.6 & 51.0 \\
% %Error&Static 6 &0.8  &0.7  &0.8  &0.9  &1.1   &1.1\\
% Error&Static 3 & 5.4 & 4.6 & 3.2 & 4.5 & 5.5  & 4.8\\
% Error&Ours     & 4.2 & 2.7 & 2.5 & 2.2 & 1.5 & 3.3 \\
% \midrule
% Acc&Static 12      &48.5  &56.9  & 70.2 &48.1  &69.8  & 64.6 \\
% %Error&Static 6 &0.6  &0.6  & 0.5  &1.0  &0.7   &0.6\\
% Error&Static 3 &4.9  &2.6  & 2.0 &4.4  &3.5   & 1.6\\
% Error&Ours     &3.6  &1.0  & 0.9 &3.6  &2.3  & 0.9 \\
% \bottomrule
% \end{tabular}
% }
% \vspace{-0.8em}
% \end{table}

\begin{table*}[t]
%\vspace{-1em}
\caption{We evaluate each classifier's score and accuracy using Static 3 and our adaptative method, comparing their 10-dimensional vectors---each element representing the score or accuracy of a specific attribute---against Static 12 using the Mean Squared Error. The evaluation is repeated three times, with average error reported, and Std in the supp. Our method provides accurate performance estimates with fewer test images and smaller errors than Static 3, optimizing the evaluation process. (Static $\#$) represents the number of test images used in each level of difficulty for each attribute of a given class.}
\label{table: compare}
\footnotesize
\centering
\begin{tabular}{l l| c c c  c c c }
\toprule
&Classifier &RN101 &ViT-B16& ConvN-B &C-RN101 &C-ViT-B16& C-ConvN-B\\
\midrule
Score&Static 12      & 35.3 & 43.8 & 59.2 & 36.4 & 57.6 & 51.0 \\
Score Error&Static 3 & 5.4 & 4.6 & 3.2 & 4.5 & 5.5  & 4.8\\
Score Error&Ours     & 4.2 & 2.7 & 2.5 & 2.2 & 1.5 & 3.3 \\
\midrule
Acc&Static 12      &48.5  &56.9  & 70.2 &48.1  &69.8  & 64.6 \\
Acc Error&Static 3 &4.9  &2.6  & 2.0 &4.4  &3.5   & 1.6\\
Acc Error&Ours     &3.6  &1.0  & 0.9 &3.6  &2.3  & 0.9 \\
\bottomrule
\end{tabular}
%\vspace{-1.8em}
\end{table*}

\subsection{Detailed error analysis}\label{sec:exp}
Since our generated dataset provides granular labels for attributes such as size, color, lighting, occlusions, and style, we can identify specific failure modes in model performance. We analyze the mistake types of the six classifiers in Table~\ref{table:attributes_results}. 
Similar to what was observed in the ImageNet validation set~\citep{idrissi2022imagenet}, models with similar overall scores tend to have similar per-attribute scores. For instance, ConvNext-Base and CLIP ViT B16 not only share similar overall scores (59.2 vs 57.6) but also have 6 out of 10 very close attribute scores. Most models, even as their overall scores improve, consistently struggle with certain factors like size, texture, style, and viewpoint. Conversely, they perform well on factors such as object position and image quality. In addition to analyzing attribute-level errors, our dataset enables a detailed difficulty-level analysis for each classifier, as shown in Tables~\ref{tab:easy_scores}, \ref{tab:medium_scores}, and \ref{tab:hard_scores} in the appendix. Across all models, the performance decreases as the difficulty level increases. Attributes like `Texture', `Style', and `Viewpoint' generally have lower accuracies, especially at the `Hard' level. One special case is for the `size' attribute; all six models perform generally well on easy and medium difficulty but face significant challenges at a hard level, which often includes examples with many tiny objects.

\section{Discussion and Limitations}

The typical way one evaluates a visual recognition model is to compute its average top-1 accuracy over a standard test set (e.g., ImageNet). However, there have been concerns that models can get over optimized on such benchmarks \citep{recht2019imagenet}. We believe that our work can help in a better assessment of models in two ways. First, since we can always generate (on-the-fly) a new fresh test dataset for evaluation, there is much less chance for a model overfitting, even if one is simply interested in classification accuracy. Second, the hierarchical learning score measures a different dimension of learning ability; it is less about how accurate a model is, but more about whether that capability is grounded in some fundamental learning principle, or in memorizing examples. 

However, this score is not immune to extreme cases. If the model to be tested has completely overfitted to test images, then it can get \{\checkmark, \checkmark, \checkmark \} for most of them and hence get a high hierarchical learning score. Finally, while our approach of using DALL-E 3 to generate a synthetic dataset offers a novel way to study model behavior across varying difficulty levels, it has certain limitations. DALL-E sometimes struggles with generating images for unfamiliar classes, such as ``African Hunting Dog,''  often producing images of other dog breeds labeled incorrectly, which can lead to noisy annotations. Additionally, for complex or uncommon prompts like ``A carousel in an amusement park, almost entirely hidden behind a festival tent,'' DALL-E may generate simplified images, such as a festival tent near a carousel but without the intended occlusion, resulting in unintended ``easy'' sample. These issues could introduce bias into our dataset, potentially affecting the balanced evaluation of models. We manually reviewed and removed some of these examples, but future work could explore more advanced generative models to address these challenges.

% \begin{table}[t]
% %\vspace{-1em}
% \caption{Score of different models for each attribute. Bold/underline indicates best/second best.}
% \vspace{-0.8em}
% \label{table:attributes_results}
% \tabcolsep=0.1cm
% \resizebox{\columnwidth}{!}{
% %\footnotesize
% \centering
% \begin{tabular}{l c c c c c c c  c c c}
% \toprule
% Attributes & Color & Light & Occlu& Pos & Quality & Rot & Size & Style & Texture & View \\
% \midrule
%  ResNet101& 41.1&31.1&40.2&39.1&47.0&34.8&30.4&23.7&30.4&35.2 \\
%  ViT-B16& 47.7 &39.7 & 46.6 &48.9 &56.9 &48.0 &36.3 &34.2 &37.8 &41.6\\
%  ConvNext-B& \textbf{63.7} &\textbf{60.6} &\textbf{59.4} &\textbf{86.7} &\underline{66.2} &\textbf{77.5} &\textbf{40.8} &41.3 &\underline{42.9} &\textbf{52.6}  \\
%  CLIP-RN101& 39.5& 33.4&33.3&62.2&36.6&39.7&34.6&30.2&23.4&31.2 \\
%  CLIP-ViT-B16& \underline{62.2}&\underline{58.9}&46.6&\underline{85.3}&\textbf{67.2}&\underline{59.8}&\underline{39.8}&\textbf{53.3}&\textbf{50.4}&\underline{52.4}  \\
%  CLIP-ConvNext-B& 53.1 &49.9 &\underline{48.4} &79.4 &57.4 &47.0 &36.7 &\underline{45.1} &41.1 &47.7\\
% % \midrule
% %  &  &  &  &  &  &   \\
% \bottomrule
% \end{tabular}
% }
% %\vspace{-1.8em}
% \end{table}

\begin{table}[t]
%\vspace{-1em}
\caption{Score of different models for each attribute. Bold/underline indicates best/second best.}
\label{table:attributes_results}
\footnotesize
\centering
\begin{tabular}{l c c c c c c c  c c c}
\toprule
Attributes & Color & Light & Occlu& Pos & Quality & Rot & Size & Style & Texture & View \\
\midrule
 ResNet101& 41.1&31.1&40.2&39.1&47.0&34.8&30.4&23.7&30.4&35.2 \\
 ViT-B16& 47.7 &39.7 & 46.6 &48.9 &56.9 &48.0 &36.3 &34.2 &37.8 &41.6\\
 ConvNext-B& \textbf{63.7} &\textbf{60.6} &\textbf{59.4} &\textbf{86.7} &\underline{66.2} &\textbf{77.5} &\textbf{40.8} &41.3 &\underline{42.9} &\textbf{52.6}  \\
 CLIP-RN101& 39.5& 33.4&33.3&62.2&36.6&39.7&34.6&30.2&23.4&31.2 \\
 CLIP-ViT-B16& \underline{62.2}&\underline{58.9}&46.6&\underline{85.3}&\textbf{67.2}&\underline{59.8}&\underline{39.8}&\textbf{53.3}&\textbf{50.4}&\underline{52.4}  \\
 CLIP-ConvNext-B& 53.1 &49.9 &\underline{48.4} &79.4 &57.4 &47.0 &36.7 &\underline{45.1} &41.1 &47.7\\
% \midrule
%  &  &  &  &  &  &   \\
\bottomrule
\end{tabular}
%\vspace{-1.8em}
\end{table}

\vspace{-5pt}
\section{Conclusion}
\vspace{-2pt}

We explored whether modern visual recognition models exhibit human-like learning behaviors when classifying images of varying difficulty. By leveraging advanced generative models, we generated a synthetic test dataset annotated with class, attribute, and difficulty level. Our findings reveal that most models do demonstrate a structured understanding of example difficulty, even without explicit supervision. This insight opens up new possibilities for model evaluation, leading us to develop an adaptive testing method that significantly reduces evaluation time. Additionally, the synthetic dataset itself, with its detailed annotations, offers a valuable resource for more granular model analysis. Our work contributes a novel perspective on understanding learning dynamics in visual recognition models and proposes an efficient, dynamic evaluation approach.

\paragraph{Acknowledgment} This work was supported in part by NSF IIS2404180, the Institute of Information \& Communications Technology Planning \& Evaluation (IITP) grant funded by the Korea government (MSIT) (No. RS2022-00187238, Development of Large Korean Language Model Technology for Efficient Pre-training), and the Microsoft Accelerate Foundation Models Research Program.

%--------------------------------------------------------------
%     Bibliography
%--------------------------------------------------------------
\bibliography{ref}
\bibliographystyle{abbrvnat}

\newpage 

%\section*{Appendix}
%\label{sec:appendix}
%\input{secs/appendix}

\clearpage

\subsection{Prompt design for image generation pipeline}~\label{sec:prompt_design}
\begin{figure}[ht]
    \centering
    \includegraphics[width=0.4\textwidth]{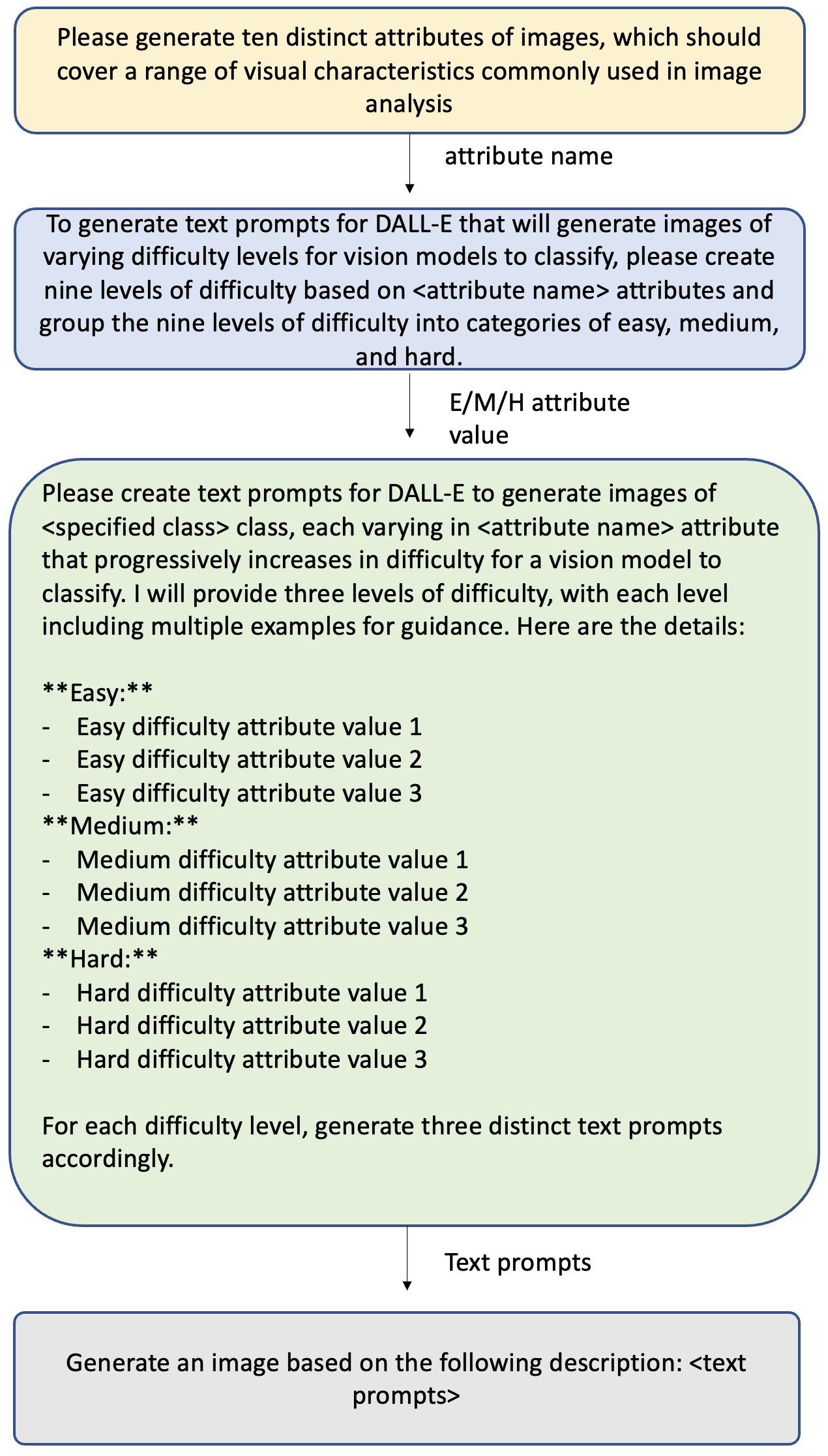}
    \caption{Desinged prompt for test set generation pipeline. }
    \label{fig:image_gen_pipeline}
\end{figure}

Please see Fig.~\ref{fig:image_gen_pipeline} for the detailed view of all the prompts used to create the final text caption used by DALLE-3 to generated the images. Note, to achieve finer granularity in difficulty design, we create nine distinct levels, allowing for a more nuanced representation of attribute variation across a spectrum. This finer resolution ensures incremental differences between levels, preventing gaps or uneven difficulty progression. In contrast, directly generating only three broad levels may oversimplify the difficulty range. As shown in the blue box of Fig.~\ref{fig:image_gen_pipeline}, we prompt GPT with``group the nine levels of difficulty into categories of easy, medium, and hard" to consolidate them into three final categories.

\vspace{-1.8em}
\subsection{User study pipeline for evaluating difficulty levels of generated images} \label{sec:user_stduy}
To assess the alignment between difficulty levels of generated images and human perception, we conducted a pairwise comparison study and analyzed the results using statistical correlation metrics. The study pipeline consists of the following steps.

\noindent\textbf{Dataset and sampling strategy} Our dataset consists of 100 classes, each with 10 attributes and three difficulty levels per attribute (Easy, Medium, Hard), with 12 images per difficulty level. To ensure feasibility for the user study, we randomly sampled a subset of 900 images: 10 classes, 10 attributes per class, 3 difficulty levels per attribute, and 3 images per difficulty level.

\begin{figure}[ht]
    \centering
    \includegraphics[width=0.6\textwidth]{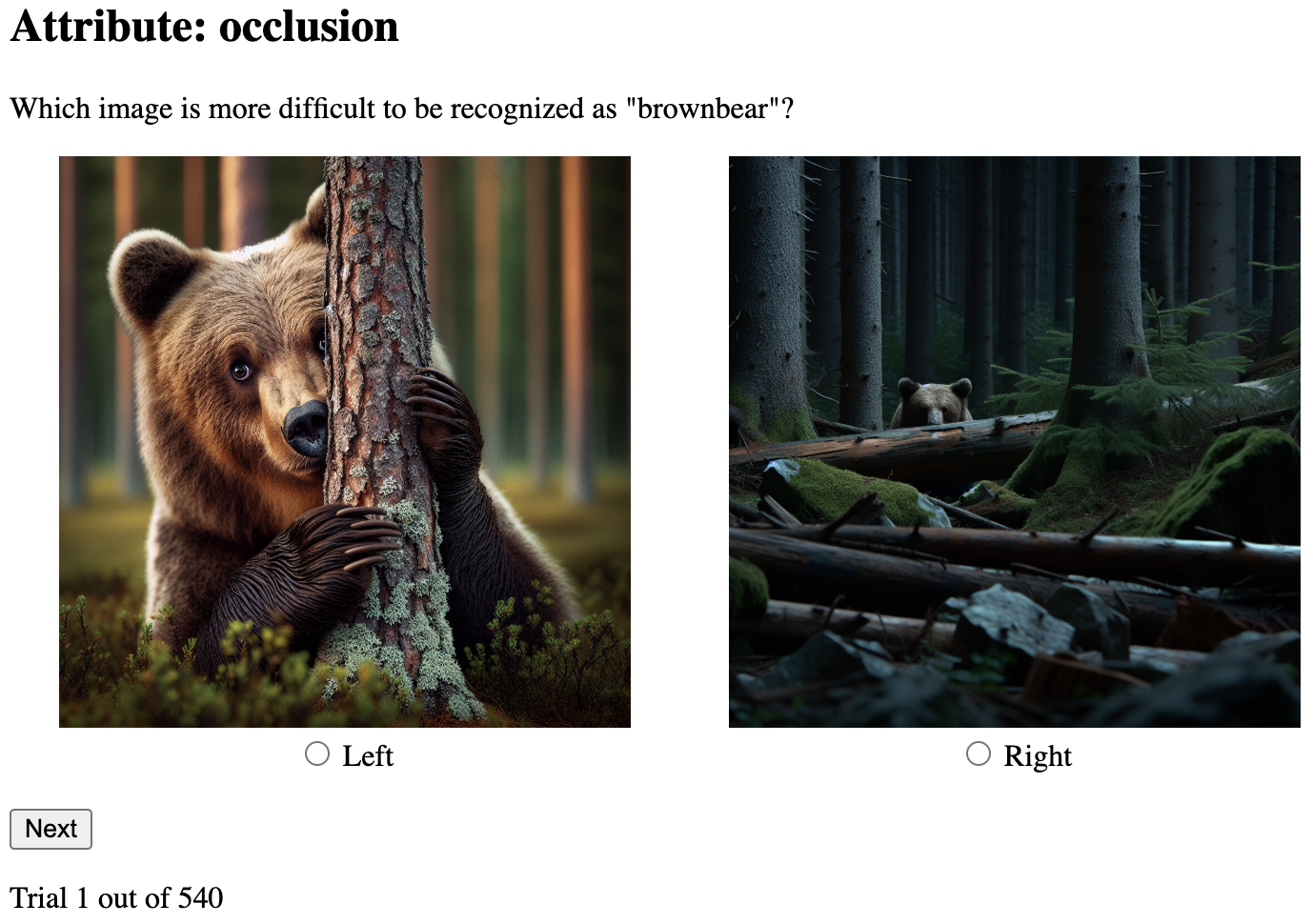}
    \caption{Interface of user study. }
    \label{fig:interface}
    \vspace{-1.0em}
\end{figure}

\noindent\textbf{Study design and pairwise comparison setup} To evaluate difficulty perception, participants compared difficulty levels within each attribute rather than across attributes or classes. Each attribute underwent three pairwise comparisons: Easy vs. Medium, Medium vs. Hard, and Easy vs. Hard. 
Participants were presented with two images at a time and asked: \texttt{``Which image is more difficult to be recognized as [class]?"}. We show the interface of user study in Fig.~\ref{fig:interface}.
Each pairwise comparison consisted of two images, one from each difficulty level. The total number of comparisons was: 27 per attribute, 270 per class, and 2700 across all classes. 

\noindent\textbf{Experimental setup}
We recruited ten participants, each evaluating two classes. Since each class contains 270 comparisons, each participant completed a total of $270 \times 2 = 540$ comparisons. This distribution ensured that all 2700 selected images were evaluated while maintaining overlap across participants to enhance robustness. To reduce bias, image placements (left vs. right) were randomized, and each image appeared in multiple comparisons.

\noindent\textbf{Analysis and ranking inference}
To derive a global difficulty ranking from human responses, we applied the Bradley-Terry Model (BTM), which estimates a continuous latent difficulty score $\lambda_i$ for each image based on pairwise comparisons. Given two images, $i$ and $j$, the probability of selecting $i$ as more difficult is:
\begin{equation}
    P(i \succ j) = \frac{e^{\lambda_i}}{e^{\lambda_i} + e^{\lambda_j}}
\end{equation}
Higher $\lambda_i$ values indicate greater perceived difficulty.

\noindent\textbf{Correlation analysis: difficulty levels of generated images vs. human-inferred difficulty}
To quantify the alignment between difficulty levels of generated images (Easy = 1, Medium = 2, Hard = 3) and human rankings, we computed Pearson correlation ($r$) to measure linear alignment, Spearman rank correlation ($\rho$) to evaluate ordinal agreement, and Kendall’s Tau ($\tau$) to assess pairwise consistency. The computed values were: $r = 0.871, \quad \rho = 0.883, \quad \tau = 0.749$.
% \begin{equation}
% \begin{aligned}
%     r &= 0.871  \\
%     \rho &= 0.883  \\
%     \tau &= 0.749  
% \end{aligned}
% \end{equation}
These results indicate a strong alignment between difficulty of generated labels and human perception.

% \noindent\textbf{Interpretation and Conclusion}
% High Pearson correlation ($r > 0.7$) confirms a strong absolute correlation between AI difficulty levels and human difficulty scores, while high Spearman rank correlation ($\rho > 0.7$) suggests that AI preserves the correct ranking order of difficulty levels. The high Kendall’s Tau ($\tau > 0.5$) indicates AI labels maintain consistent pairwise ordering with human perception. The strong correlations validate AI-generated difficulty classification, though minor discrepancies highlight areas for refinement. This study provides a quantitative framework for evaluating difficulty classification models and emphasizes the importance of human-centered validation in AI-generated difficulty assessments.

\subsection{Ablation of hyperparameters for the GRE testing round}\label{sec:hyper}
The effectiveness of the adaptive test in approximating overall performance should not be highly sensitive to minor changes in test structure. To validate this, we conducted an ablation study on test parameters by modifying two aspects: (i) the number of questions in the first and second rounds and (ii) the distribution of second-round questions based on first-round performance. 

In this modified setting, referred to as \textit{ours new}, the model receives five questions in the first round, consisting of \textbf{1 easy, 3 medium, and 1 hard} question, ensuring an average difficulty of medium. The score range is from \textbf{0 to 11}. The second-round question distribution, in terms of easy (E), medium (M), and hard (H) questions, is adjusted as follows:

\begin{itemize}
    \item \textbf{Score = 0} \quad (E=4, M=0, H=0)
    \item \textbf{Score = [1,3]} \quad (E=3, M=1, H=0)
    \item \textbf{Score = [4,6]} \quad (E=1, M=2, H=1)
    \item \textbf{Score = [7,10]} \quad (E=0, M=1, H=3)
    \item \textbf{Score = 11} \quad (E=0, M=0, H=4)
\end{itemize}

Using this revised test format, we report model accuracy below, consistent with the results presented in Table~\ref{table:ablation_of_hyper}. For comparison, we replicate the results of the existing adaptive test (\textit{ours old}) alongside the overall accuracy of the static 12-question test. The results indicate that performance remains largely consistent with the previous test version.

\begin{table}[t]
\caption{Impact of different hyperparameters for the GRE testing round. The results of different hyperparameters remain largely consistent.}
\label{table:ablation_of_hyper}
\footnotesize
\centering
\begin{tabular}{l c c c }
\toprule
Model & Acc (Ours Old) & Acc (Ours New) & Acc (All)  \\
\midrule
 ConvNext-B & 70.5 & 70.8 & 70.2\\
 ViT-B16       & 56.8 & 57.4 & 56.9 \\
 ResNet101     & 48.5 &3.89 & 3.07 \\
 CLIP ConvNext-B& 64.8& 64.1 & 64.6  \\
 CLIP ViT-B16& 70.4 & 69.2 & 69.8 \\
 CLIP ResNet101& 48.0 & 48.8 & 48.1\\
\bottomrule
\end{tabular}
%\vspace{-1.8em}
\end{table}

\subsection{Visualizing the Difficulty of Test Samples}
\begin{figure*}[t]
    \centering
    %\vspace{-0.7cm}
    % \includegraphics[width=1\textwidth,page=2]{figures/tesear.pdf}
    \includegraphics[width=0.9\textwidth]{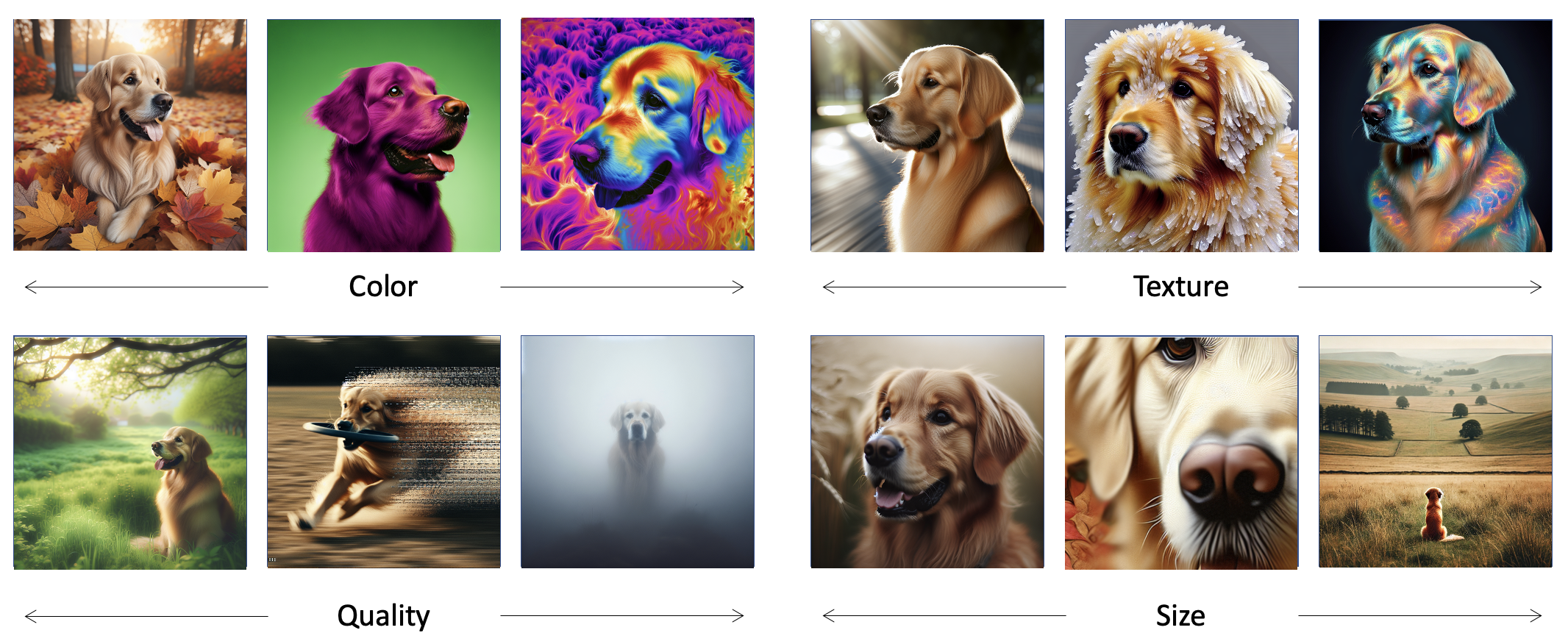}
    \vspace{-0.3cm}
    \caption{\textbf{Visualizing the difficulty of test samples.} All of the images are generated using our proposed pipeline. In each quadrant, we focus on one attribute (e.g., color, in the top left), and from left to right we show the images becoming progressively more difficult to be classified correctly.}
    \label{fig:difficulty_examples_more}
    \vspace{-0.1in}
\end{figure*}

\begin{figure*}[t]
    \centering
    \vspace{-0.7cm}
    \includegraphics[width=1\textwidth]{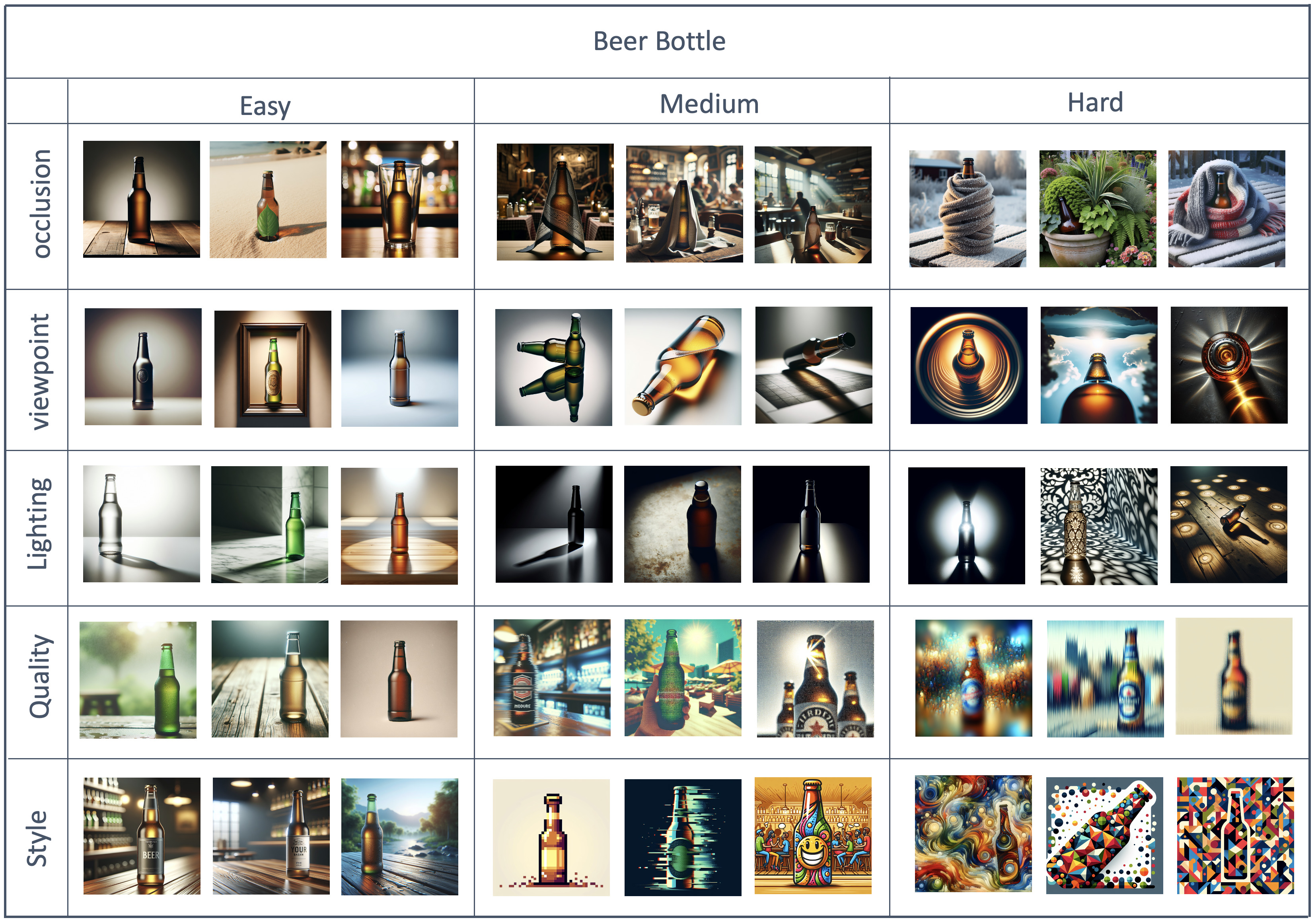}
    \vspace{-0.5cm}
    \caption{Visualizing the class of Beer Bottle.}
    \label{fig:samples1}
    \vspace{-0.1in}
\end{figure*}

\begin{figure*}[t]
    \centering
    \vspace{-0.7cm}
    \includegraphics[width=1\textwidth]{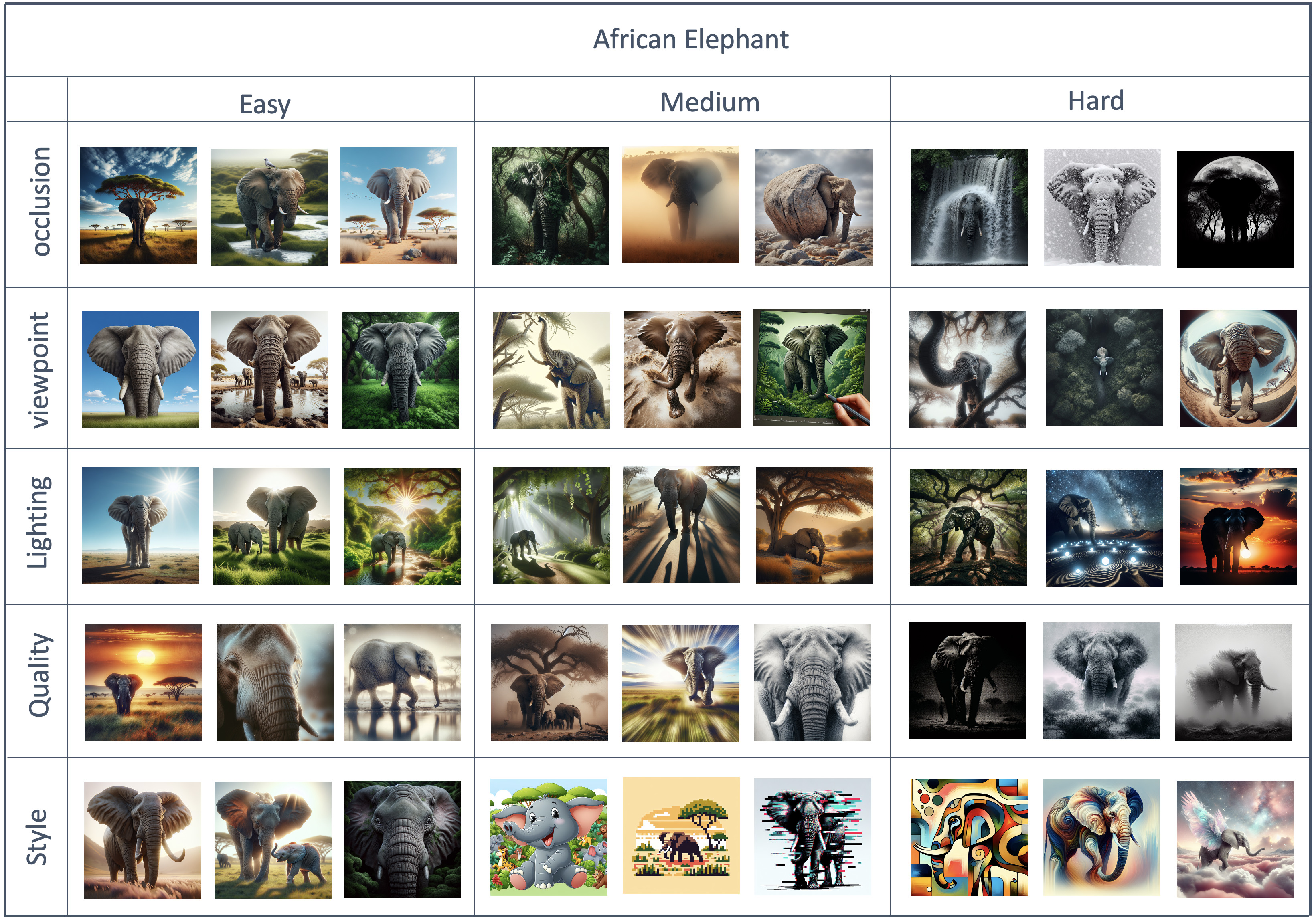}
    \vspace{-0.5cm}
    \caption{Visualizing the class of African elephant.}
    \label{fig:samples2}
    \vspace{-0.1in}
\end{figure*}

\begin{figure*}[t]
    \centering
    \vspace{-0.7cm}
    \includegraphics[width=1\textwidth]{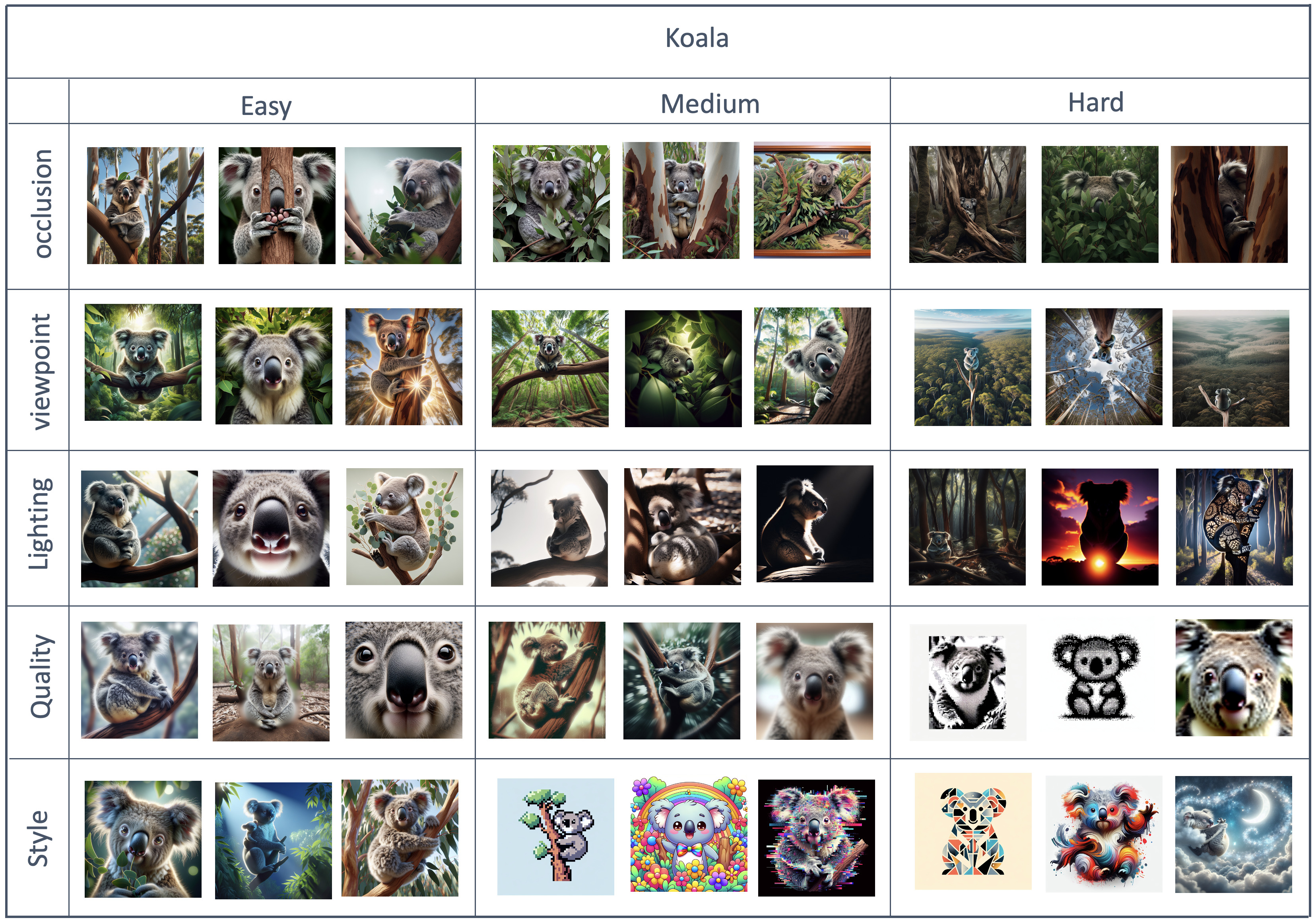}
    \vspace{-0.5cm}
    \caption{Visualizing the class of Koala.}
    \label{fig:samples3}
    \vspace{-0.1in}
\end{figure*}

\begin{figure*}[t]
    \centering
    \vspace{-0.7cm}
    \includegraphics[width=1\textwidth]{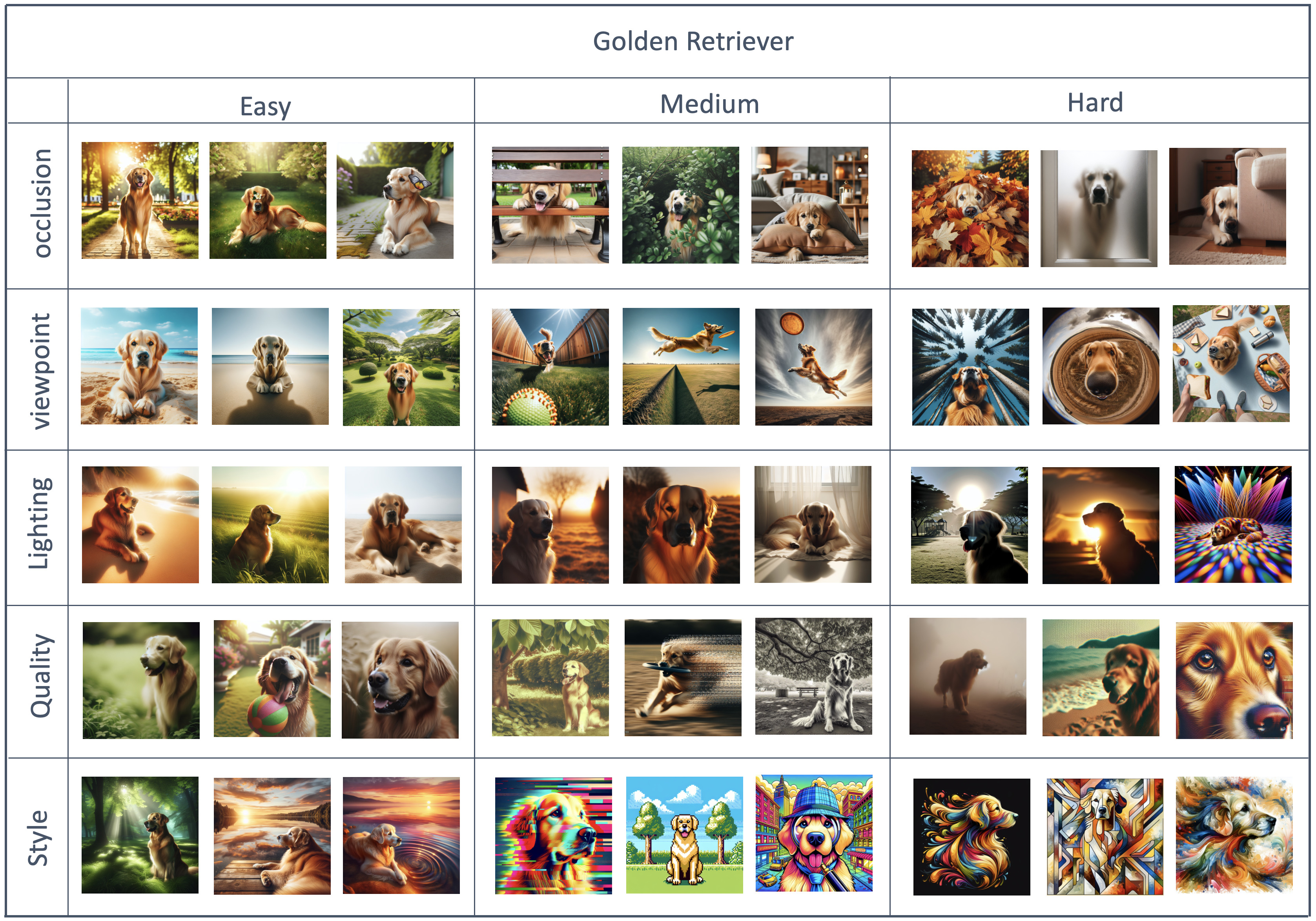}
    \vspace{-0.5cm}
    \caption{Visualizing the class of golden retriever.}
    \label{fig:samples4}
    \vspace{-0.1in}
\end{figure*}

We present additional images featuring a golden retriever as the main subject, focusing on attributes such as color, texture, quality, and size. From left to right, the images are arranged to become progressively more challenging for accurate classification. Please see Fig.~\ref{fig:difficulty_examples_more}. Finally, we also show more examples for other classes along with their attributes in Fig.~\ref{fig:samples1}, \ref{fig:samples2}, \ref{fig:samples3}, \ref{fig:samples4}.

\subsection{Detailed Error Analysis}

In addition to analyzing attribute-level errors, our generated dataset enables a detailed difficulty-level analysis for each classifier, as shown in Tables~\ref{tab:easy_scores}, Table~\ref{tab:medium_scores}, and Table~\ref{tab:hard_scores}.
Across all models, the performance decreases as the difficulty level increases. This is a general trend for each attribute, indicating that all models struggle more with "Hard" samples compared to "Easy" and "Medium" ones. Additionally, attributes like "Texture," "Style," and "Viewpoint" generally have lower accuracies, especially at the "Hard" level. This suggests that these attributes pose more significant challenges for current deep-learning models.

\begin{table*}[ht]
    \centering
    \begin{adjustbox}{max width=\textwidth}
    \begin{tabular}{lccccccc}
        \toprule
        \textbf{Attribute} & \textbf{CLIP ResNet101} & \textbf{ResNet101} & \textbf{CLIP ViT B16} & \textbf{ViT B16} & \textbf{CLIP ConvNext Base} & \textbf{ConvNext Base} & \textbf{Average (Attributes)} \\
        \midrule
        Color     & 58.89 & 70.74 & \underline{83.33} & 75.56 & \textbf{84.07} & 83.70 & 76.38 \\
        Lighting  & 67.04 & 67.04 & \textbf{91.11} & 77.41 & \underline{87.41} & 82.59 & 78.77 \\
        Occlusion & 65.93 & 76.67 & \underline{84.81} & 80.00 & \textbf{88.52} & 86.67 & 80.77 \\
        Position  & 97.78 & \underline{96.67} & \textbf{100.00} & 97.04 & 99.26 & 97.04 & 97.96 \\
        Quality   & 69.26 & 78.52 & \textbf{89.26} & \underline{80.74} & 87.41 & 87.41 & 82.77 \\
        Rotate    & \underline{99.26} & 96.67 & \textbf{100.00} & 97.78 & \textbf{100.00} & 99.26 & 98.49 \\
        Size      & 98.52 & 97.04 & \textbf{100.00} & \underline{98.15} & \textbf{100.00} & 99.26 & 98.83 \\
        Style     & 71.48 & 68.89 & \underline{82.96} & 78.52 & \textbf{85.56} & 82.22 & 78.27 \\
        Texture   & 42.96 & \underline{56.67} & \textbf{77.78} & 67.04 & 75.19 & 75.19 & 65.64 \\
        Viewpoint & 63.70 & \underline{77.41} & 86.67 & 84.81 & 84.44 & \textbf{89.63} & 81.11 \\
        \midrule
        \textbf{Average} & 73.08 & 77.13 & \textbf{89.59} & 83.00 & \underline{89.19} & 88.30 & \\
        \bottomrule
    \end{tabular}
    \end{adjustbox}
    \caption{Accuracy for different attributes at the easy difficulty level. Bold indicates the highest score, and underline denotes the second highest. The rightmost column shows the average accuracy of each attribute.}
    \label{tab:easy_scores}
\end{table*}

\begin{table*}[ht]
    \centering
    \begin{adjustbox}{max width=\textwidth}
    \begin{tabular}{lccccccc}
        \toprule
        \textbf{Attribute} & \textbf{CLIP ResNet101} & \textbf{ResNet101} & \textbf{CLIP ViT B16} & \textbf{ViT B16} & \textbf{CLIP ConvNext Base} & \textbf{ConvNext Base} & \textbf{Average (Attributes)} \\
        \midrule
        Color     & 50.37 & 51.48 & \underline{78.89} & 66.29 & 69.63 & \textbf{81.85} & 66.42 \\
        Lighting  & 48.52 & 47.78 & \textbf{84.44} & 55.93 & \underline{75.19} & 80.37 & 65.71 \\
        Occlusion & 47.41 & 57.78 & \underline{72.59} & 62.96 & 71.48 & \textbf{80.00} & 65.37 \\
        Position  & 67.41 & 38.89 & \underline{93.70} & 54.44 & 91.11 & \textbf{94.81} & 73.73 \\
        Quality   & 43.70 & 60.74 & \underline{78.89} & 67.78 & 75.19 & \textbf{77.04} & 67.22 \\
        Rotate    & 56.67 & 44.44 & \underline{94.07} & 69.63 & 75.19 & \textbf{96.30} & 72.05 \\
        Size      & 62.22 & 54.07 & \underline{81.85} & 70.74 & \textbf{85.19} & 85.19 & 73.54 \\
        Style     & 49.26 & 35.19 & \textbf{84.44} & 56.67 & \underline{78.52} & 66.29 & 61.06 \\
        Texture   & 40.37 & 49.26 & \textbf{78.89} & 57.41 & \underline{69.26} & 68.52 & 60.62 \\
        Viewpoint & 44.07 & 56.29 & \underline{80.74} & 65.56 & 67.78 & \textbf{82.96} & 66.23 \\
        \midrule
        \textbf{Average} & 50.40 & 49.69 & \textbf{82.85} & 62.44 & 75.65 & \underline{81.03} & \\
        \bottomrule
    \end{tabular}
    \end{adjustbox}
    \caption{Accuracy for different attributes at the medium difficulty level. Bold indicates the highest score, and underline denotes the second highest. The rightmost column shows the average accuracy of each attribute.}
    \label{tab:medium_scores}
\end{table*}

\begin{table*}[ht]
    \centering
    \begin{adjustbox}{max width=\textwidth}
    \begin{tabular}{lccccccc}
        \toprule
        \textbf{Attribute} & \textbf{CLIP ResNet101} & \textbf{ResNet101} & \textbf{CLIP ViT B16} & \textbf{ViT B16} & \textbf{CLIP ConvNext Base} & \textbf{ConvNext Base} & \textbf{Average (Attributes)} \\
        \midrule
        Color     & 29.26 & 28.52 & \underline{48.52} & 31.48 & 37.04 & \textbf{47.04} & 36.98 \\
        Lighting  & 17.41 & 13.70 & \underline{38.15} & 22.22 & 30.74 & \textbf{45.19} & 27.57 \\
        Occlusion & 18.15 & 22.22 & 24.07 & \underline{30.00} & 26.67 & \textbf{44.07} & 27.53 \\
        Position  & 50.74 & 38.89 & \underline{77.41} & 34.07 & 68.15 & \textbf{80.37} & 58.27 \\
        Quality   & 24.81 & 32.22 & \underline{55.93} & 45.56 & 43.33 & \textbf{52.59} & 42.07 \\
        Rotate    & 16.30 & 14.44 & \underline{32.59} & 24.81 & 19.63 & \textbf{62.59} & 28.06 \\
        Size      & 4.81  & 1.85  & \underline{3.70}  & 3.70  & 1.85  & \textbf{4.44} & 3.39 \\
        Style     & 10.37 & 6.67  & \underline{30.37} & 11.85 & 20.37 & \textbf{21.48} & 16.52 \\
        Texture   & 10.00 & 14.44 & \underline{29.26} & 20.74 & 18.52 & \textbf{22.22} & 19.20 \\
        Viewpoint & 16.67 & 14.07 & \textbf{29.63} & 18.89 & \underline{28.51} & 28.15 & 22.26 \\
        \midrule
        \textbf{Average} & 19.65 & 18.60 & \underline{36.36} & 26.83 & 29.75 & \textbf{38.82} & \\
        \bottomrule
    \end{tabular}
    \end{adjustbox}
    \caption{Accuracy for different attributes at the hard difficulty level. Bold indicates the highest score, and underline denotes the second highest. The rightmost column shows the average accuracy of each attribute.}
    \label{tab:hard_scores}
\end{table*}

\subsection{Hierarchical Learning Score of Additional Models}
As Section 3.2 mentions Hierarchical Learning Score (HLS), we include an additional six classifiers: ResNet 18, ResNet 50, ConvNext Large, ConvNext Small, ViT Small 16, and ViT Large 16. Their Hierarchical Learning Scores are provided in Table~\ref{table:hscore_perc}.

\begin{table*}[ht]
%\vspace{-1em}
\caption{Hierarchical Learning Score of additional six visual recognition models.}
\label{table:hscore_perc}
\scriptsize
\centering
\begin{tabular}{l c c c  c c c }
\toprule
Classifer &ResNet18 &ResNet50& ConvNext-L &ConvNext-S &ViT-S16& ViT-L16\\
 HLS &  86.52 & 86.19 &90.56 &  88.22  & 85.00 &  85.04 \\
\bottomrule
\end{tabular}
%\vspace{-1.8em}
\end{table*}

\subsection{Standard deviation across multiple runs}\label{sec:std}
We ran our experiments shown in Table 2 three times. The standard deviation of classification scores, both for ours vs static 3 baseline, is shown in Table~\ref{table:std}. We see that the results are consistent (low standard deviation) for all the models.

\begin{table}[ht]
\caption{The standard deviation of classification scores for ours vs static 3 baseline.}
\label{table:std}
\scriptsize
\centering
\begin{tabular}{l c c c  c c c }
\toprule
Classifer &ConvNeXt B &ViT B16	& RN101 &CLIP ConvNeXt B &CLIP ViT B16& CLIP RN101\\
 Std of Score Error (ours) &  1.3 & 1.9 & 1.1 &  0.7  & 1.0 &  0.8 \\
 Std of Score Error (Static3) &  1.9 & 2.3 & 2.1 &  1.4  & 1.6 &  1.5 \\
\bottomrule
\end{tabular}
\end{table}

\subsection{More confidence visualization for the Easy, Medium, and Hard difficulty}
\begin{figure}[ht]
    \centering
    \subfloat[Classification confidence for ViT-B16 model.]{
        \includegraphics[width=0.48\textwidth]{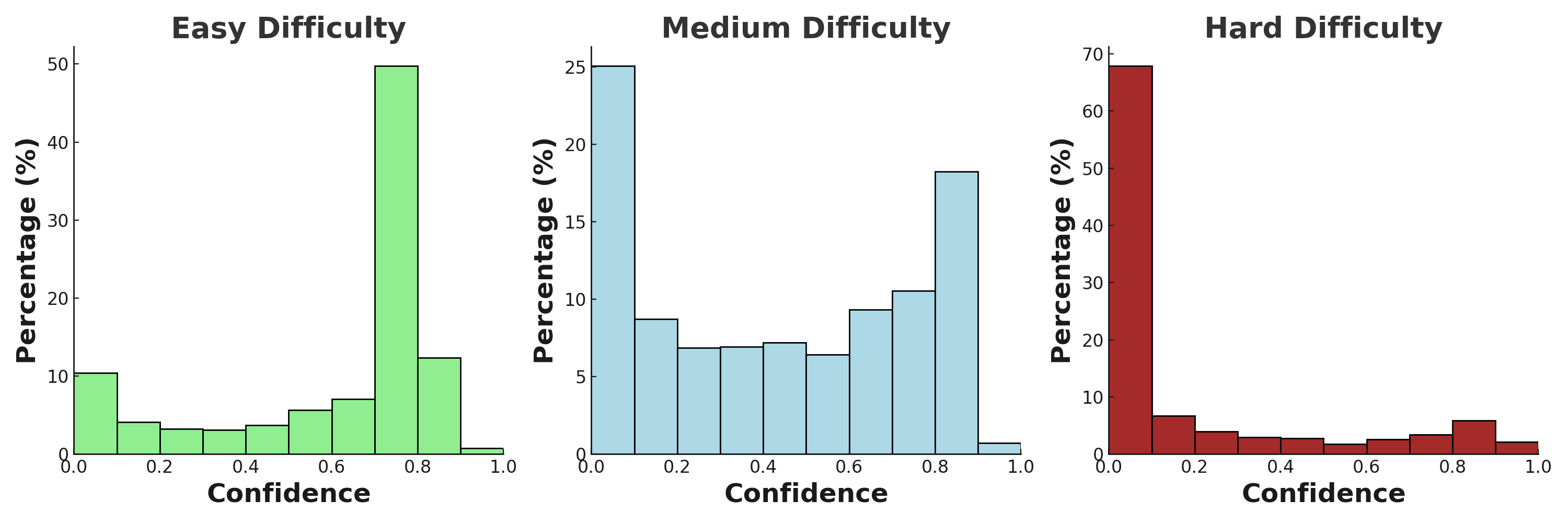}
        \label{fig:vitb16_acc}
    }
    \hfill
    \subfloat[Classification confidence for CLIP-ViT-B16 model.]{
        \includegraphics[width=0.48\textwidth]{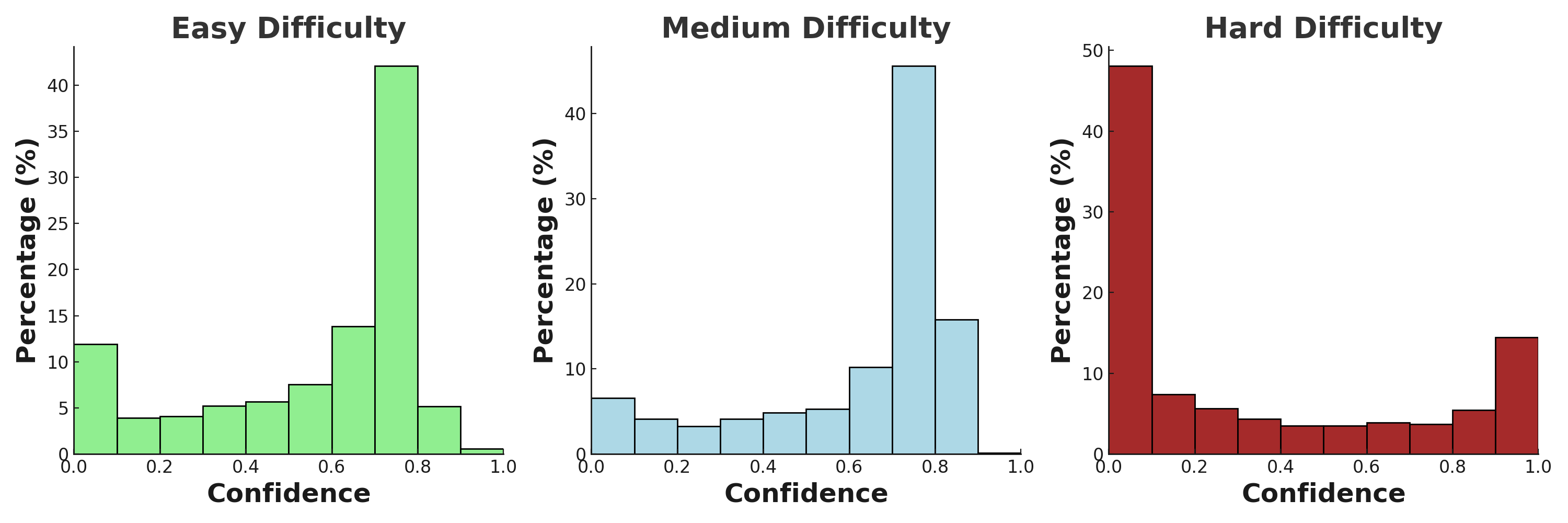}
        \label{fig:clip_vitb16_acc}
    }
    \caption{Comparison of classification confidence: (a) ViT-B16 and (b) CLIP-ViT-B16.}
    \label{fig:comparison}
\end{figure}

In this section, we visualize the distribution of prediction confidence across the difficulty levels for several classifiers, using our generated dataset. Please see Fig.~\ref{fig:vitb16_acc} and \ref{fig:clip_vitb16_acc}. We see that they follow a similar trend as described in Fig.~\ref{fig:synthetic}, where the distribution of confidence is progressively decreasing as we move from easy $\rightarrow$ hard samples.

% \begin{figure}[ht]
%     \centering
%     \includegraphics[width=0.48\textwidth]{vitb16.png}
%     \caption{Classification confidence for ViT-B16 model. }
%     \label{fig:vitb16_acc}
% \end{figure}

% \begin{figure}[ht]
%     \centering
%     \includegraphics[width=0.48\textwidth]{clip_vitb16.png}
%     \caption{Classification confidence for CLIP-ViT-B16 model.}
%     \label{fig:clip_vitb16_acc}
% \end{figure}

\subsection{Attribute of Variation Definitions}\label{appendix:attributes}

\noindent\textbf{Position:} The location or placement of the object within the frame of the image. It can indicate whether the object is centered, towards the edge, or even partially out of view. 

\noindent\textbf{Viewpoint:} Describes the angle or perspective from which the object is observed, such as front, side, top-down, or oblique view. The viewpoint affects the amount of detail visible and can reveal or obscure specific features of the object.

\noindent\textbf{Quality:} Indicates the overall clarity and resolution of the image. High-quality images have fine details and little noise, while low-quality images may appear blurry, pixelated, or noisy, making it harder to discern specific features.

\noindent\textbf{Rotate:} Describes the orientation of the object in the image. An object can be upright, tilted, or flipped. The rotation can affect the perception and recognition of the object's standard appearance. 

\noindent\textbf{Occlusion:} Occurs when parts of the main object are blocked or obscured by other objects in the scene. This can make it challenging to identify the full structure of the object. 

\noindent\textbf{Size:} Refers to the object's scale within the image. Size can be influenced by the object's actual size, its distance from the camera, or the zoom level. 

\noindent\textbf{Lighting:} Lighting in the image is either brighter or darker when compared to the prototypical images.

\noindent\textbf{Color:} Color can indicate the object's natural appearance, the time of day, or the overall mood. 

\noindent\textbf{Texture:} Refers to the surface quality or pattern seen on the object, such as smooth, rough, glossy, or matte. 

\noindent\textbf{Style:} Indicates the visual aesthetics or artistic rendering of the image. This could include photographic styles (e.g., realistic, abstract, cartoonish), drawing styles, or filters applied to the image.

\subsection{List of 100 Object Categories}\label{appendix:categories}

We selected 100 object categories from the 1,000 classes in ImageNet for our study. These categories represent a diverse range of items, animals, and objects, including:
Objects: catamaran, wooden spoon, hourglass, stopwatch, iPod, plate, crate, turnstile, frying pan, comic book, pencil box, cash machine, school bus, obelisk, volleyball, lifeboat, computer keyboard, CD player.
Animals: malamute, koala, goose, meerkat, gazelle, bullfrog, loggerhead turtle, box turtle, iguana, Komodo dragon, rock python, diamondback rattlesnake, scorpion, wolf spider, black grouse, flamingo, king penguin, killer whale, Chihuahua, Maltese dog, beagle, Afghan hound, Irish wolfhound, Border collie, Rottweiler, Bernese mountain dog, Dalmatian, Siberian husky, lion, tiger, American black bear, ladybug, fire salamander, hummingbird, goldfinch, toucan, peacock, lobster, Dungeness crab, zebra, bison, hippopotamus, giraffe, kangaroo, platypus, woodpecker, raccoon, skunk, bat, otter, seahorse, jellyfish, sea anemone, coral, stork, crane, tortoise, parrot.
Food-related: beer bottle, lipstick, mixing bowl, mashed potato.
Others: cliff, black widow, lakeside, sock, great white shark, ostrich, bald eagle, vulture, American alligator, African elephant, golden retriever.
This wide range of categories ensures a comprehensive evaluation of model performance across various domains.

\end{document}